\documentclass[journal]{IEEEtran}
\ifCLASSINFOpdf
  \usepackage[pdftex]{graphicx}
\else
\fi
\usepackage{amsmath}
\usepackage{amsfonts}
\usepackage{romannum}
\usepackage{algorithm}
\usepackage{algpseudocode}

\usepackage{multirow}
\usepackage{booktabs}

\usepackage{url}

\usepackage{booktabs}
\usepackage{xspace}
\usepackage{pifont}
\usepackage{hyperref}
\usepackage{adjustbox}
\usepackage{tcolorbox}

\newcommand{\method}{UniTrans\xspace}
\newcommand{\data}{$\mathcal{P}_d$\xspace} 
\newcommand{\task}{$\mathcal{P}_t$\xspace}
\newcommand{\myparagraph}[1]{\textbf{#1}\xspace}

\begin{document}
%
\title{UniTrans: A Unified Vertical Federated Knowledge Transfer Framework for Enhancing Cross-Hospital Collaboration}
%
%
%

\author{Chung-ju~Huang,
        Yuanpeng~He,
        Xiao~Han,
        Wenpin Jiao,
        Zhi Jin,~\IEEEmembership{Fellow,~IEEE,}
        and~Leye~Wang,~\IEEEmembership{Member,~IEEE}
\thanks{Chung-ju~Huang, Yuanpeng~He, Wenpin~Jiao, Zhi~Jin, and Leye~Wang are with the Key Lab of High Confidence Software Technologies, Ministry of Education, Beijing 100816, China, and also with the School of Computer Science, Peking University, Beijing 100871, China (e-mail: chongruhuang.pku@gmail.com; heyuanpeng@stu.pku.edu.cn; jwp@pku.edu.cn; zhijin@pku.edu.cn; leyewang@pku.edu.cn).}
\thanks{Xiao Han is with School of Economics and Management, Beihang University, Beijing 100191, China (e-mail: xh\_bh@buaa.edu.cn).}
\thanks{Manuscript received April 19, 2005; revised August 26, 2015.}}

%
%

\markboth{Journal of \LaTeX\ Class Files,~Vol.~14, No.~8, August~2015}%
{Shell \MakeLowercase{\textit{et al.}}: Bare Demo of IEEEtran.cls for IEEE Journals}
%



\maketitle

\begin{abstract}
Cross-hospital collaboration has the potential to address disparities in medical resources across different regions. However, strict privacy regulations prohibit the direct sharing of sensitive patient information between hospitals. Vertical federated learning (VFL) offers a novel privacy-preserving machine learning paradigm that maximizes data utility across multiple hospitals. Traditional VFL methods, however, primarily benefit patients with overlapping data, leaving vulnerable non-overlapping patients without guaranteed improvements in medical prediction services. While some knowledge transfer techniques can enhance the prediction performance for non-overlapping patients, they fall short in addressing scenarios where overlapping and non-overlapping patients belong to different domains, resulting in challenges such as feature heterogeneity and label heterogeneity. To address these issues, we propose a novel unified vertical federated knowledge transfer framework (Unitrans). Our framework consists of three key steps. First, we extract the federated representation of overlapping patients by employing an effective vertical federated representation learning method to model multi-party joint features online. Next, each hospital learns a local knowledge transfer module offline, enabling the transfer of knowledge from the federated representation of overlapping patients to the enriched representation of local non-overlapping patients in a domain-adaptive manner. Finally, hospitals utilize these enriched local representations to enhance performance across various downstream medical prediction tasks. Experiments on real-world medical datasets validate the framework's dual effectiveness in both intra-domain and cross-domain knowledge transfer. The code of \method is available at \url{https://github.com/Chung-ju/Unitrans}.
\end{abstract}


%
\IEEEpeerreviewmaketitle

\section{Introduction}
\label{sec:intro}
%
%
%
%

\IEEEPARstart{F}{ederated} Learning (FL) \cite{mcmahan2017communication} is a distributed machine learning paradigm that has garnered significant attention for its ability to facilitate collaborative model training while preserving data privacy. In FL, multiple parties independently train models on their local data, sharing only model parameters with a central server that aggregates them into a global model. This global model typically achieves better performance than the individual local models. Unlike centralized machine learning, FL ensures that the raw data from each party remains private and is not exchanged. FL has been widely adopted by major technology companies, such as Google and Apple, where it is used to enhance services like Face ID, Siri, and predictive text in Gboard \cite{yang2018applied}.

\begin{figure}[ht]
  \centering
  \includegraphics[width=\linewidth]{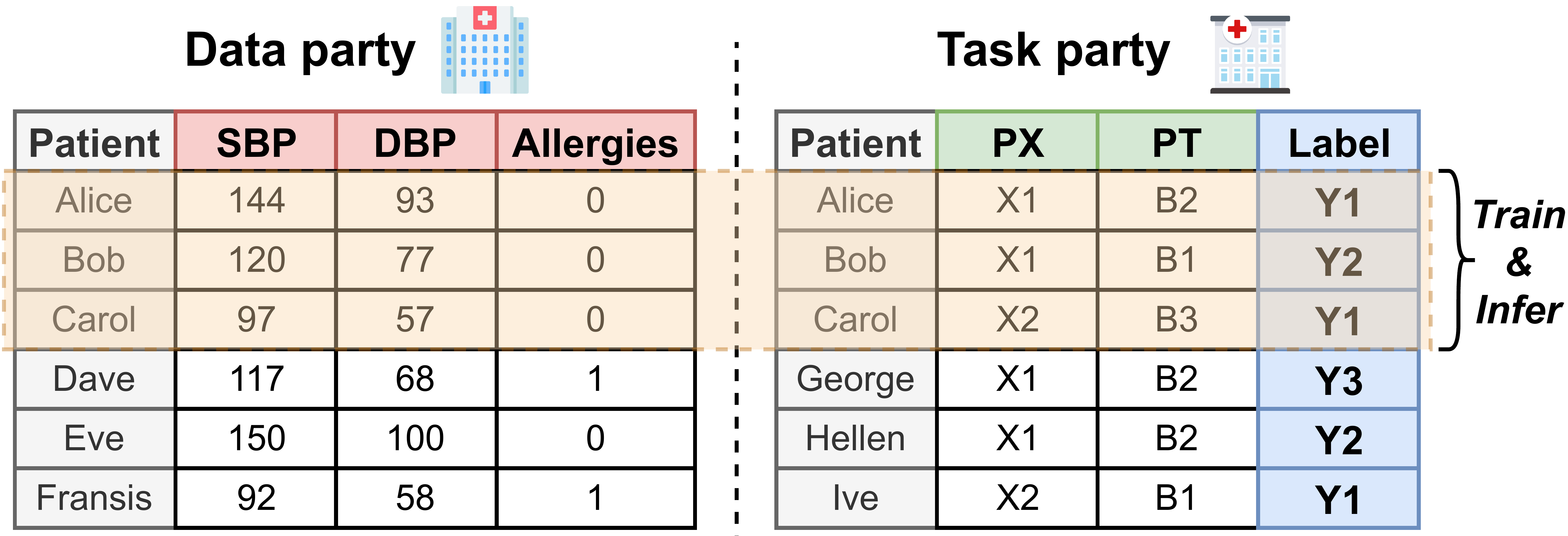}
  \caption{Healthcare application scenario of VFL.}
  \label{fig:vfl}
\end{figure}

In general, FL can be categorized into two types: Horizontal FL (HFL) and Vertical FL (VFL). HFL was first proposed by Google in their seminal work~\cite{mcmahan2017communication} and is applicable to scenarios where different parties have data with similar feature spaces but different user spaces. The clients in HFL are usually various terminal devices. Each client first trains a local model using isolated data and then sends the model parameters to the server. The server aggregates the model parameters using methods such as FedAvg~\cite{mcmahan2017communication} and then sends the global model back to each client. In contrast, VFL operates in situations where different parties, often organizations, hold distinct feature spaces of the same set of users. 
The stringent data privacy requirements imposed by regulations such as the Health Insurance Portability and Accountability Act (HIPAA)~\cite{act1996health} and the EU's General Data Protection Regulation (GDPR)~\cite{voigt2017eu} have significantly constrained multi-institutional research. In this context, VFL has emerged as a pivotal technology for enabling secure and effective cross-party medical collaboration~\cite{HuangW023,CheKPSLCH22}. 
Therefore, we take hospital collaboration as scenario, as illustrated in Fig.~\ref{fig:vfl}. 
The VFL task initiator who holds the labels is referred to as the task party \task. The data party \data, who provide abundant data and collaborate in training/inference, supports the task party. \task aims to collaborate with \data to provide enhanced healthcare prediction services for overlapping patients.

\textbf{Motivation}. The successful adoption of current VFL methods is highly dependent on how many overlapping samples exist between parties. Hence, most VFL collaborations are conducted by involving at least one giant \data with abundant data. For instance, FDN (federated data network) \cite{LiuFCXY21} includes anonymous data from one of the largest social network service providers in China and thus can cover most user samples from other data holders (e.g., customers of banks). However, this makes giant data holders occupy a dominant position over other small data holders in VFL, which could lead to unfair trades and data monopoly in the digital economy.\footnote{\url{https://www.theguardian.com/technology/2015/apr/19/google-dominates-search-real-problem-monopoly-data}} 
In the scenario depicted in Fig.~\ref{fig:vfl}, \task is typically a small hospital in remote area, while \data is a large hospital with abundant features. The application of VFL by \task and \data for disease prediction on overlapping patients raises two issues: 
\ding{172} The characteristics and attributes of the data from \task are often overshadowed by large \data with more extensive features. This dominance hampers the \task's ability to leverage the specificity of local data to provide more personalized treatment for patients. 
\ding{173} Patients need to visit multiple hospitals to access better federated services. However, patients restricted to remote hospitals due to geographical or socioeconomic factors may face unequal access to medical resources. 
Thus, it is essential to focus on and protect this vulnerable group. 
To alleviate this defect and expand the application scenarios of VFL, \textit{a VFL-based collaborative framework that can benefit local non-overlapping samples is urgently needed.}

\textbf{Status Quo and Challenges.}
With the advancement of VFL, there has been a growing body of work focusing on the usability of non-overlapping samples~\cite{HuangW023,RenYC22,KangLL22,fedhssl,LiuKXCY20}. Their common approach involves training on cross-party overlapping samples and transferring federated knowledge to their local non-overlapping samples. 
However, in real scenarios, especially in healthcare, knowledge transfer is not simplistic. Because non-overlapping patients and overlapping patients may come from different domains, such as from different departments of \task. 
This will lead to two key challenges~\cite{sciadv.adp6040,YangYFZYLH16}: feature heterogeneity and label heterogeneity. 
\begin{itemize}
    \item \textit{Feature heterogeneity}~\cite{ZhouTPT19,EbrahimiCZC23,ZhiHZG23}. The overlap between \task and \data may mainly involve patient data from internal medicine departments, which usually include patients with chronic diseases (such as hypertension, diabetes, etc.)~\cite{TianWSAHKLFW24}. However, non-overlapping patients may come from patient data in surgical departments, involving diseases that require surgical intervention (such as tumors, trauma, etc.) as well as pre- and postoperative evaluations, anesthesia information, postoperative recovery, and other related data.
    \item \textit{Label heterogeneity}~\cite{KouwL21,YuL23,HeQKCTZ23,Azizzadenesheli22}. The label definitions for internal medicine and surgery may be different, resulting in deviations in the model's understanding and use of predicted labels after knowledge transfer~\cite{ZhouBZZF22}. Internal medicine labels may be long-term labels about disease management, such as ``well-controlled hypertension'', ``long-term stable diabetes'', etc. Surgical labels might be about surgical risk, acute changes in medical condition, or trauma management, such as ``high-risk surgery'', ``good postoperative recovery'', or ``acute heart failure''~\cite{ZhouJWXT22}.
\end{itemize}

\begin{figure}[htbp]
  \centering
  \includegraphics[width=\linewidth]{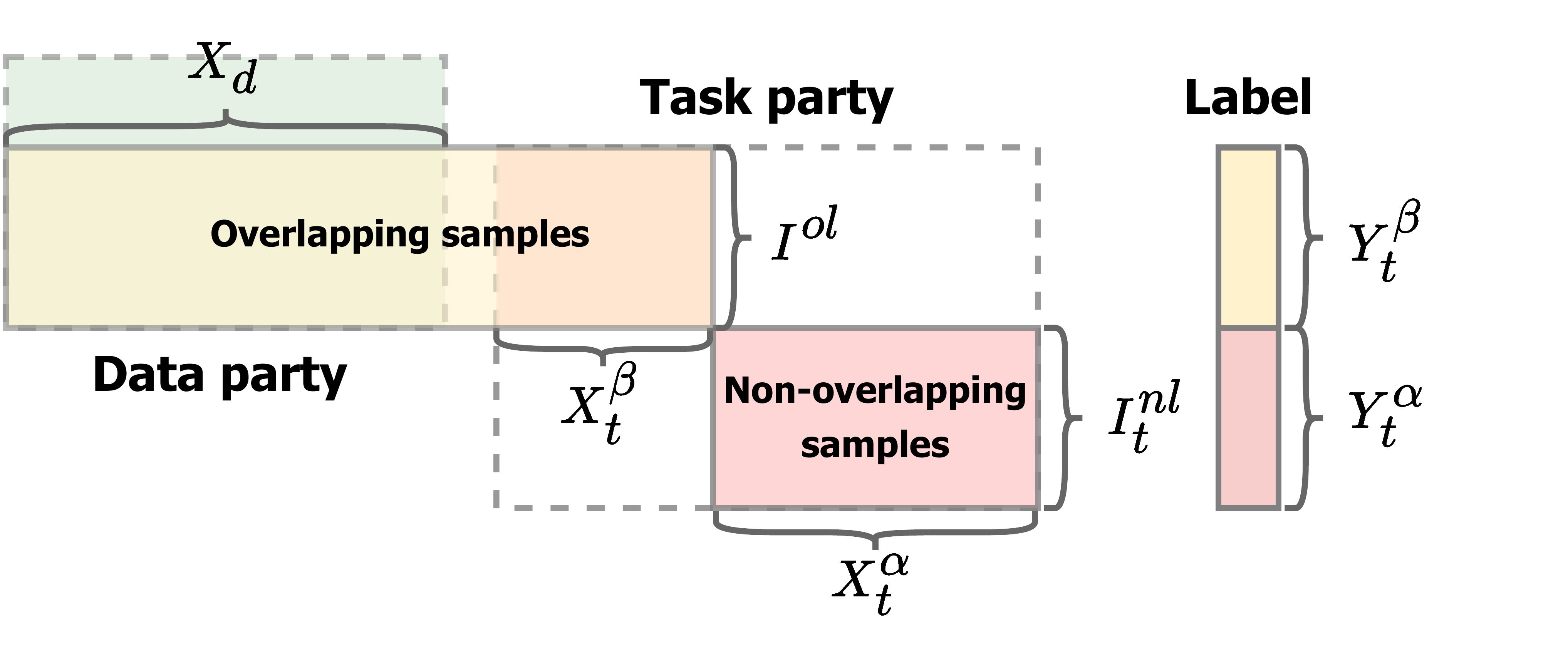}
  \caption{Two challenges in VFL knowledge transfer.}
  \label{fig:challenge}
\end{figure}

These challenges are illustrated in Fig.~\ref{fig:challenge}. 
Although the medical and surgical patient populations differ in disease types and treatments, they often share some underlying physiological characteristics and clinical indicators, which may be highly predictive in tasks such as diagnosis and risk assessment~\cite{YuZLLXD22,kraft2023domain}. By transferring knowledge from overlapping patients' data from internal medicine, surgical models can be helped to capture these invariant features, thereby improving the predictive accuracy of non-overlapping patients. If the model learns how to identify the association between hypertension and other complications in internal medicine, and transfers this knowledge to surgery, the model will be able to better predict surgical risks or postoperative complications in surgical patients. 
We have summarized and compared existing knowledge transfer techniques in Tbl.~\ref{tab:compare} (details in Sec~\ref{subsec:vkt}), revealing that current approaches are insufficient to simultaneously address both challenges. 
Therefore, there is an urgent need to develop advanced vertical federated knowledge transfer technologies that can effectively handle both intra-domain and cross-domain scenarios, addressing the challenges posed by complex real-world medical tasks.

\textbf{Our work.}
We propose a unified framework for vertical federated knowledge transfer (\method) that enables the transfer of medical knowledge from overlapping samples across collaborative healthcare networks to each hospital's local, non-overlapping samples. 
The key challenge is addressing the issue of feature and label heterogeneity during knowledge transfer. Specifically, overlapping and non-overlapping samples may originate from different domains, requiring effective methods to handle these discrepancies. 
We begin by employing novel unsupervised federated representation extraction techniques, such as federated singular vector decomposition (FedSVD) \cite{chai2022practical}, to derive federated latent representations from overlapping samples. 
Next, we introduce a cross-attention-based local knowledge transfer module, designed to effectively distill knowledge from these multi-party federated latent representations into local non-overlapping samples. To enhance knowledge transfer, we incorporate cross-domain adaptation by maximizing the mutual information between the source and target domains, which ensures effective alignment of features across domains. 
To minimize redundancy in knowledge transfer during multi-party collaboration (involving more than two parties), we incorporate the concept of contrastive learning. This approach enhances the uniqueness of the knowledge acquired through one-to-one learning, ensuring more efficient and targeted knowledge transfer. 
Finally, each participating hospital can leverage the joint representation of overlapping samples as a guide to enrich the feature representations of its local samples through the trained knowledge transfer module. 
Our proposed knowledge transfer mechanism offers several key advantages:

\begin{itemize}
  \item \textit{Knowledge transfer to local samples}. Unlike most VFL methods that focus solely on overlapping samples, our proposed mechanism is designed to enhance the prediction performance of local non-overlapping samples across different parties through vertical knowledge transfer. This approach enables hospitals with a limited number of overlapping samples to derive substantial benefits from collaboration, ensuring effective collaboration even in scenarios with sparse overlapping data.

  \item \textit{Domain-adaptive transfer}. The feature alignment techniques and task-independent transfer mechanisms integrated into \method ensure strong performance in both intra-domain and cross-domain knowledge transfer scenarios. This capability not only enhances the adaptability of the model but also broadens its applicability to diverse real-world scenarios, where feature heterogeneity and label heterogeneity are common challenges.
  
  \item \textit{Scalable to multiple hospitals}. Our mechanism achieves efficient multi-party learning with computational complexity that scales linearly with the number of participating hospitals. Simultaneously, it guarantees effective knowledge transfer during multi-party collaboration, leading to enhanced prediction performance.
\end{itemize}

\textbf{Contribution}. In summary, this work makes the following contributions:

\begin{enumerate}
  \item To the best of our knowledge, this work is the first one to explore how to enable vertical knowledge transfer from overlapping samples to each hospital's non-overlapping samples in a domain-adaptive manner.

  \item We propose a novel domain-adaptive vertical-knowledge-transfer framework, \method, to transfer medical knowledge from overlapping samples to local samples. \method demonstrates a dual capability in knowledge transfer, making it highly versatile and impactful in both intra-domain and cross-domain applications.

  \item Experiments conducted on seven real-world medical datasets validate the effectiveness of our mechanism for knowledge transfer and the generalizability of the enriched feature representations for local samples. These results highlight how \method empowers hospitals with limited medical resources to deliver improved healthcare services through VFL collaboration.
\end{enumerate}

\section{Related Work}

\subsection{Vertical Knowledge Transfer}
\label{subsec:vkt}
Existing VFL mechanisms~\cite{XuZZWZLL24,10745553,LiuKZPHYOZY24} primarily focus on improving the prediction performance of overlapping samples. In contrast, this work focuses on transferring knowledge from overlapping samples to enhance the prediction performance of each party's local non-overlapping samples.

We summarize the prior knowledge transfer techniques in Tbl.~\ref{tab:compare}. 
FTL trains feature extractors for \task and \data on overlapping samples, mapping heterogeneous feature spaces to a common latent subspace~\cite{LiuKXCY20}. FTL relies on source domain labels during training. Therefore, while FTL can address feature heterogeneity, it cannot handle label heterogeneity. 
VFL-Infer~\cite{RenYC22} transfers knowledge from the federated model on overlapping samples to local models by extracting soft labels generated by the federated model. 
FedCVT~\cite{KangLL22} utilizes existing representations to estimate the missing feature representations for \task and predicts pseudo-labels for unlabeled samples to construct an extended training set. Both VFL-Infer and FedCVT are incapable of handling scenarios with both feature and label heterogeneity. 
VFedTrans~\cite{HuangW023} first uses FedSVD~\cite{chai2022practical} on overlapping samples to obtain federated latent representations, and then trains a single auto-encoder on overlapping and non-overlapping samples to perform representation distillation. This makes VFedTrans unable to handle the feature heterogeneity problem of overlapping and non-overlapping samples. In the knowledge transfer of VFedTrans, only overlapping samples can obtain the distillation loss constraint of the federated latent representation. This further reduces the effectiveness of knowledge transfer for non-overlapping samples.
FedHSSL~\cite{fedhssl} first uses cross-party self-supervised learning (SSL) to train a cross-party encoder on overlapping samples, and then uses the cross-party encoder to train a local encoder on local samples (including overlapping and non-overlapping).
Although VFedTrans and FedHSSL can address label heterogeneity, they still fail to resolve feature heterogeneity. Overall, existing vertical knowledge transfer techniques are unable to simultaneously address feature heterogeneity and label heterogeneity. The proposed \method overcomes this limitation, enabling the improvement of prediction performance for non-overlapping samples in more complex scenarios.

\begin{table}[htbp]
\centering
\caption{Comparison of existing knowledge transfer techniques in VFL.}
\begin{tabular}{ccc}
    \toprule
    \ding{61}\textbf{Method} & \textbf{Feature Heterogeneity} & \textbf{Label Heterogeneity} \\ \midrule
    VFL-Infer~\cite{RenYC22}      & No  & No   \\ 
    FedCVT~\cite{KangLL22}        & No  & No   \\ 
    FedHSSL~\cite{fedhssl}        & No  & Yes  \\ 
    VFedTrans~\cite{HuangW023}    & No  & Yes  \\ 
    FTL~\cite{LiuKXCY20}          & Yes & No   \\ 
    \textbf{Ours}                 & Yes & Yes  \\ 
    \bottomrule
\end{tabular}
\label{tab:compare}
\end{table}

\subsection{Federated Domain Learning}
Existing federated domain learning~\cite{GuanLPLX24,YuanMCWLK23,10772332} primarily focuses on improving the prediction performance of the target domain in HFL scenarios. Only a small number of studies~\cite{kang2022privacy,YanWHHZXLZ24} have focused on the domain learning problem in VFL. PrADA~\cite{kang2022privacy} proposes a vertical federated adversarial domain adaptation method aimed at learning high-order features for multiple domains on overlapping samples across multiple participants, thereby improving the interpretability of predictions. Fed-CRFD~\cite{YanWHHZXLZ24} investigates domain shift issues caused by different modalities on overlapping samples to facilitate MRI reconstruction. These studies focus solely on domain learning on overlapping samples. In contrast, our work aims to address the cross-domain challenges in knowledge transfer from overlapping to non-overlapping samples.

\section{Problem Formulation}
\label{sec:problem}

In this section, we clarify the definitions of key concepts used in this paper. Afterward, we formulate our research problem. The used notations can be found in Appendix.

\subsection{Concepts}
Our approach enables non-overlapping patients from all hospitals to benefit from the collaboration. VFL typically involves a two-party scenario~\cite{PangYSW23}, consisting of one task hospital \task and one data hospital \data.

\begin{itemize}
    \item \textit{Task Hospital}. A task hospital \task has a set of patients with features $X_t$ and a task label $Y_t$ to predict. The patient IDs of \task are denoted as $I_t$. 
    \item \textit{Data Hospital}. A data hospital \data has a set of samples with features $X_{d}$. The data hospital's sample IDs are denoted as $I_{d}$.
\end{itemize}
Patient IDs overlapping between \task and \data are referred to as $I_t^{ol}=I_d^{ol}=I_t \cap I_{d}$, while non-overlapping IDs of \task are referred to as $I_t^{nl}=I_t\setminus I_t^{ol}$.


\subsection{Research Problem}
\label{subsec:problem}

Given a task hospital \task and a data hospital \data, holding overlapping patient data $H_t^{ol}\in \mathbb{R}^{I_t^{ol}\times X_t^{\beta}}$ and $H_d^{ol}\in \mathbb{R}^{I_d^{ol}\times X_d}$ respectively. \task also holds the target non-overlapping patient data $H_t^{nl}\in \mathbb{R}^{I_t^{nl}\times X_t^{\alpha}}$. 
The feature domains $X_t^{\beta}$ and $X_t^{\alpha}$ and the label domains $Y_t^{nl}$ and $Y_t^{ol}$ may be heterogeneous. 
The objective is to design a knowledge transfer mechanism to predict the task label $Y_t^{nl}$ of \task's non-overlapping patients data $H_t^{nl}$ as accurately as possible.

\textbf{Remark}. Traditional VFL frameworks often require that $I_t = I_d$. However, our vertical federated knowledge transfer setting only needs that $I_t^{ol} \ne \emptyset$. Without loss of generality, we focus on validating the impact of knowledge transfer on \task to demonstrate that \method has robust service support for hospitals with limited knowledge. Specifically, the goal of \method is to enhance the task performance of \task's local $H_t^{nl}$ by transferring the knowledge from $H_t^{ol}$ and $H_d^{ol}$. This significantly broadens the practical applicability of VFL in real-world scenarios.

\section{\method's Design}
\label{sec:machanism}

\begin{figure*}
  \centering
  \includegraphics[width=\linewidth]{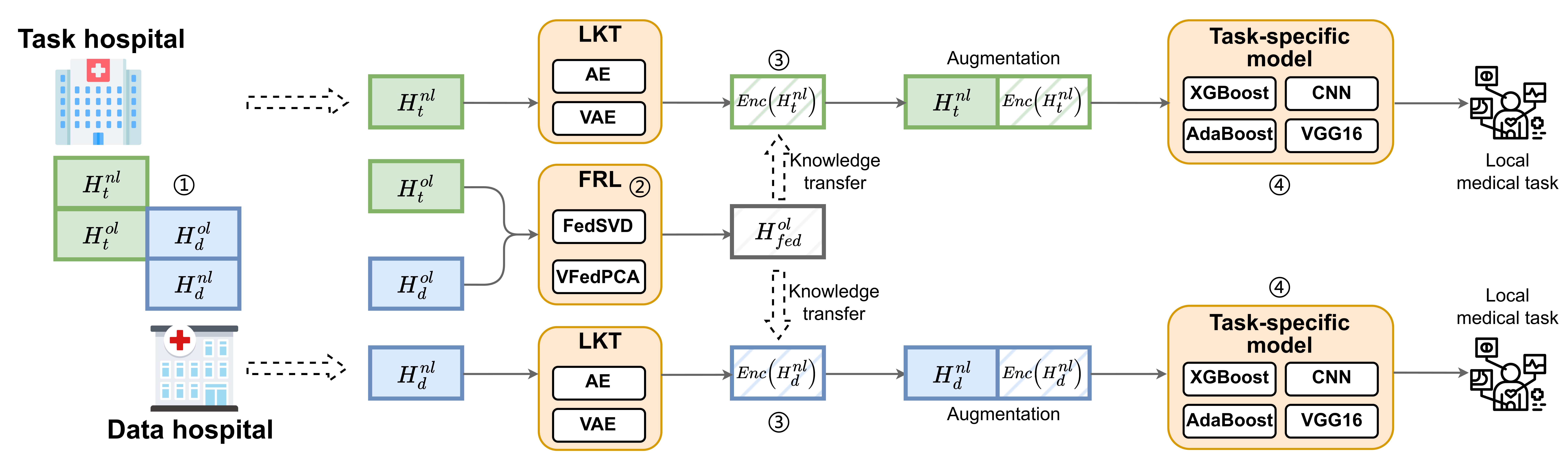}
  \vspace{-2em}
  \caption{
  Overview of \method for two hospitals. Both \task and \data can benefit from \method. 
  \ding{172} \task and \data collaborate to identify overlapping samples $H_t^{ol}$ and $H_d^{ol}$ through PSI. 
  \ding{173} With the assistance of a trusted third-party server, \task and \data conduct online FRL to generate the federated latent representation $H_{fed}^{ol}$, for the overlapping samples.
  \ding{174} Using $H_{fed}^{ol}$, both \task and \data perform LKT offline to transfer knowledge to their respective local, non-overlapping samples, $H_t^{nl}$ and $H_d^{nl}$. This step augments their local features.
  \ding{175} With the augmented features, \task and \data can carry out their respective downstream prediction tasks according to their individual requirements.
  }
  \label{fig:overview}
\end{figure*}

\subsection{Overview}

\begin{figure}[ht]
  \centering
  \includegraphics[width=\linewidth]{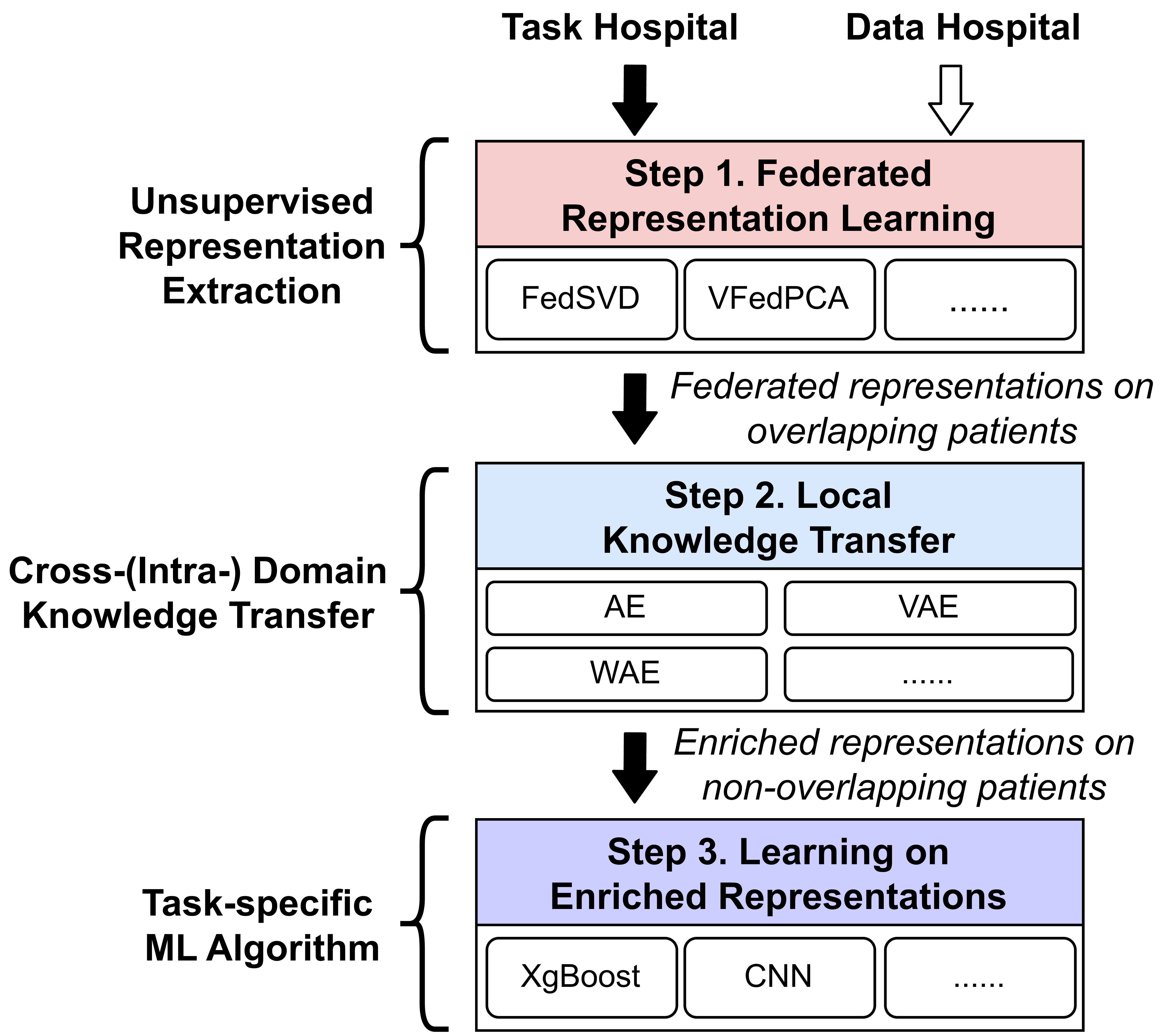}
  \caption{Overview of \method.}
  \label{fig:mechanism_overview}
\end{figure}

We demonstrate the overall process of \method (Fig.~\ref{fig:overview}). Note that before our mechanism runs, we suppose that overlapping IDs $I_t^{ol}$ between \task and \data are known, which can be learned by private set intersection (PSI) methods~\cite{Kamara2014ScalingPS}. Our framework can be simplified into three main steps (Fig.~\ref{fig:mechanism_overview}).

\begin{itemize}
  \item \textbf{Step 1. Federated Representation Learning (FRL)}. 
  First, \task and \data collaboratively learn federated latent representations $H_{fed}^{ol} \in \mathbb{R}^{|I_t^{ol}|\times |X_t^{\beta}+X_d|}$ using secure VFL techniques. In brief, these federated latent representations would incorporate the hidden knowledge among multiple parties while not leaking these parties' raw data.
	
  \item \textbf{Step 2. Local Knowledge Transfer (LKT)}. 
  Second, \task trains a domain-adaptive representation transfer module designed to transfer knowledge from federated latent representations, enriching the feature representations of local non-overlapping patients.

  \item \textbf{Step 3. Task-Specific Learning}. 
  After Step 2, the LKT module is ready for local feature augmentation. With any given label $Y_t^{nl}$ to predict, \task can leverage the enriched representations of local samples (i.e., original local features combined with the augmented representations) to perform training and inference using state-of-the-art (SOTA) machine learning (ML) algorithms.
\end{itemize}

Step 3 generally follows traditional supervised learning methods to train a task-specific prediction model, allowing the application of various ML algorithms, such as XGBoost~\cite{Chen2016XGBoostAS} and VGG16~\cite{SimonyanZ14a}. 
Next, we illustrate more details about Step 1 and 2. For simplicity, we begin by considering a scenario with a single \task and a single \data, which aligns with the common setup in VFL~\cite{PangYSW23,LiuKZPHYOZY24}. At the end of Sec.~\ref{subsec:lkt}, we will expand our discussion to address scenarios involving multiple data hospitals.

\subsection{Federated Representation Learning}
\label{subsec:frl}
The objective of Step 1 is to extract federated latent representations $H_{fed}^{ol}$ by leveraging $H_t^{ol}$ and $H_d^{ol}$ in a privacy-friendly manner. 
This step can utilize a range of vertical federated representation learning techniques to achieve effective feature extraction. 
In this work, we adopt a matrix decomposition-based federated representation method, as previous studies have demonstrated its effectiveness in extracting meaningful latent representations for ML tasks~\cite{kosinski2013private}. 
Specifically, we utilize two state-of-the-art federated matrix decomposition methods: FedSVD~\cite{chai2022practical} and VFedPCA~\cite{cheung2022vertical}. 
Below, we detail how these methods are modified and applied to learn $H_{fed}^{ol}$ by incorporating $H_t^{ol}$ and $H_d^{ol}$.

\subsubsection{FedSVD}
In FedSVD \cite{chai2022practical}, \task and \data use two random orthogonal matrices to transform the $H_t^{ol}$ and $H_d^{ol}$ into masked $\hat{H}_t^{ol}$ and $\hat{H}_d^{ol}$, respectively. This maintains the invariance of the decomposition results despite the masking transformation of the local samples. $\hat{H}_t^{ol}$ and $\hat{H}_d^{ol}$ are then uploaded to a third-party server, which applies the SVD algorithm to the masked patient samples from all hospitals. Finally, \task can reconstruct the federated latent representation based on the decomposition results.

Suppose \task holds the overlapping samples' feature matrix $H_t^{ol} \in \mathbb{R}^{|I_t^{ol}|\times|X_t^{\beta}|}$, and \data holds the overlapping samples' feature matrix $H_d^{ol} \in \mathbb{R}^{|I_d^{ol}|\times|X_d|}$ ($I_t^{ol} = I_d^{ol} = I_t \cap I_d $ is the overlapping sample ID set). Denote $H^{ol} = [H_t^{ol}|H_d^{ol}]$ (combination of both task and data hospitals' feature matrices), we want to leverage $H^{ol} = U\Sigma V^T$ (SVD) to learn the latent representations $U$,
Inspired by FedSVD \cite{chai2022practical}, we use a randomized masking method to learn $U$ as:
\begin{enumerate}
  \item A trusted key generator generates two randomized orthogonal matrices $A\in \mathbb{R}^{|I_t^{ol}|\times|I_t^{ol}|}$ and $B \in \mathbb{R}^{|X_{fed}^{ol}|\times|X_{fed}^{ol}|}$ ($|X_{fed}^{ol}| = |X_t^{ol}| + |X_d^{ol}|$). $B$ is further partitioned to two parts $B_t \in \mathbb{R}^{|X_t^{ol}|\times|X_{fed}^{ol}|}$ and $B_d \in \mathbb{R}^{|X_d^{ol}|\times|X_{fed}^{ol}|}$, i.e., $B^T = [B_t^T | B_d^T]$.
	
  \item $A$ and $B_t$ are sent to \task; $A$ and $B_d$ are sent to \data. Each hospital does a local computation by masking its own original feature matrices with the received matrices:
  \begin{equation}
    \hat{H}_k^{ol} = A H_k^{ol} B_k, \forall k \in \{t,d\}
  \end{equation}

  \item \task and \data send $\hat{H}_t^{ol}$ and $\hat{H}_d^{ol}$ respectively to a third-party server\footnote{The third-party server needs to be semi-honest. Note that in FL, such a security configuration (i.e., the information aggregation server is semi-honest) is widely accepted \cite{yang2019federated}.} and the third-party server runs SVD on the combined data matrix $\hat{H}^{ol} = \hat{U} \Sigma \hat{V}^T$, where $\hat{H}^{ol} = [\hat{H}_t^{ol}|\hat{H}_d^{ol}]$.
  $\hat{U}$ is then sent back to \task.
	
  \item \task can recover the federated latent representation of shared samples, denoted as $H_{fed}^{ol}$, by
  \begin{equation}
    H_{fed}^{ol} = U = A^T \hat U
  \end{equation}
\end{enumerate}

Compared to the original FedSVD which aims to recover both $U$ and $V$~\cite{chai2022practical}, we only need to recover $U$. Hence, in \method, only $\hat{U}$ is transmitted to \task to reduce the communication cost. The correctness of the above process depends on the fact that $H^{ol}$ and $\hat{H}^{ol}$ (multiplying $H^{ol}$ by two orthogonal matrices) must hold the same singular value $\Sigma$~\cite{chai2022practical}. 

\subsubsection{VFedPCA}
To enhance the generality of \method, we also use vertical federated principal component analysis (VFedPCA)~\cite{cheung2022vertical} to extract latent representations. Under VFedPCA's setting, each hospital makes its own federated eigenvector $u$ converge to global eigenvector $u_G$ without needing to know the mutual data of all hospitals. Each hospital is able to train the local eigenvector using local power iteration~\cite{saad2011numerical}. Then the eigenvectors from each hospital are merged into the federated eigenvector $u$. Finally, \task can use $u$ to reconstruct the original data to obtain the federated latent representation.

Suppose \task holds $H_t^{ol} \in \mathbb{R}^{|I_t^{ol}|\times|X_t^{\beta}|}$, and \data holds $H_d^{ol} \in \mathbb{R}^{|I_d^{ol}|\times|X_d|}$. Denote $H^{ol} = [H_t^{ol}|H_d^{ol}]$

\begin{enumerate}
    \item For each hospital $k \in \{t,d\}$, we calculate the largest eigenvalue $A_k=\frac{1}{|X_k|}{(H_k^{ol})}^T H_k^{ol}$ and a non-zero vector $a_k$ corresponding to the eigenvector $\delta_k(A_ka_k=\delta_k a_k)$. The number of local iterations is $L$, each hospital will compute locally until convergence as follows:
    \begin{equation}
        a_k^{l}=\frac{A_k a_k^{l-1}}{||A_k a_k^{l-1}||}, \quad \delta_k^l=\frac{A_k{(a_i^l)}^T a_k^l}{{(a_k^l)}^T a_k^l}
    \end{equation}
    where $l = 1, 2, \cdots, L$.
    \item Then each hospital upload the eigenvector $a_k^L$ and the eigenvalue $\delta_k^L$ to third-party server. The server aggregates the results and generates the federated eigenvalue:
    \begin{equation}
        u = \sum_{k\in \{t,d\}}w_k a_t^L,\quad w_k = \frac{\delta_k^L}{\sum_{k \in \{t, d\}}\delta_k^L}
    \end{equation}
    \item \task can use the federated eigenvalue $u$ to reach the $H_{fed}^{ol}$:
    \begin{equation}
        H_{fed}^{ol} = H_t^{ol}\frac{MM^T}{||MM^T||},M={(H_t^{ol})}^T u
    \end{equation}
\end{enumerate}

\subsection{Local Knowledge Transfer}
\label{subsec:lkt}

After obtaining $H_{fed}^{ol}$ for overlapping samples, Step 2 aims to enrich the \task's non-overlapping samples' representations. The overlapping and non-overlapping samples of \task may come from different domains, resulting in feature heterogeneity or label heterogeneity. However, existing knowledge transfer techniques in VFL are unable to address this challenge. 
Some studies have demonstrated that cross-attention mechanisms~\cite{VaswaniSPUJGKP17} can facilitate better feature alignment in domain adaptation~\cite{XuCWWL022,XieT024} and domain generalization~\cite{MengLCYSWZSXP22,ZhouLQXL23}. 
Building on these concepts, we developed a domain-adaptive unified knowledge transfer module utilizing a cross-attention mechanism. This module is designed with two primary objectives: \ding{172} fuse invariant features between source domain and target domain in the same subspace, and \ding{173} perform label-free knowledge transfer. 

For the patients' data $H_t^{nl}$, we use an auto-encoder $Enc$-$Dec$ to extract representations $Enc(H_t^{nl})$. We then employ a cross-attention mechanism to learn domain-invariant features $Z_t^{nl}$: 
\begin{align}
    Z_t^{nl}=attn(Enc(H_t^{nl}), H_{fed}^{ol}&\otimes \Phi) \\ \notag
    =softmax(\frac{Enc(H_t^{nl})\otimes{(H_{fed}^{ol}\otimes \Phi)}^T}{\sqrt{X_t^{nl}}})&\otimes (H_{fed}^{ol}\otimes \Phi)
\end{align}
where $\Phi$ is a learnable parameter matrix of dimension $X_{fed}^{ol}\times X_t^{nl}$ to capture $\beta$-to-$\alpha$ cross domain transformation. In this way, we transform the learned global federated knowledge into local knowledge.

\begin{figure}[ht]
  \centering
  \includegraphics[width=\linewidth]{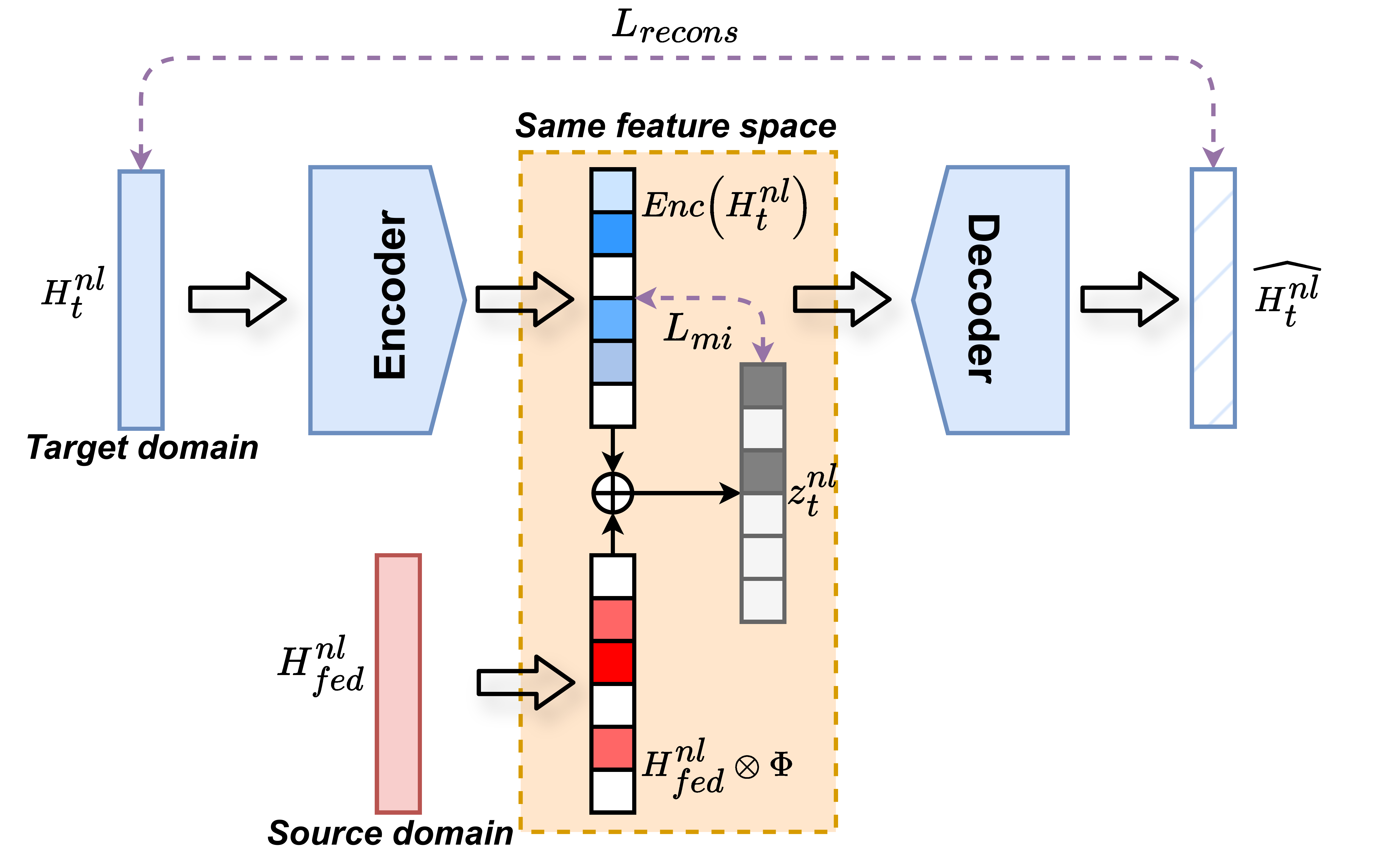}
  \caption{The process of LKT of \method.}
  \label{fig:lkt}
\end{figure}

To enhance knowledge transfer, we employ a Mutual Information Neural Estimator (MINE)~\cite{BelghaziBROBHC18} to maximize the mutual information between the target domain representations $P=Enc(H_t^{nl})$ and the transferred domain-invariant representations $Q=Z_t^{nl}$. To avoid integral computation, we use Monte-Carlo integration to estimate mutual information:
\begin{equation}
    MI(P,Q)=\frac{1}{N}\sum_{i=1}^N I(p,q,\theta)-\log(\frac{1}{N}\sum_{i=1}^N e^{I(p,q',\theta)})
\end{equation}
where $(p, q)$ is sampled from the joint distribution, $q'$ is sampled from the marginal distribution, and $I$ is a neural network with parameters $\theta$ used to estimate the mutual information between $P$ and $Q$. We incorporate the mutual information $\mathcal{L}_{mi}=MI(P,Q)$ estimated by MINE into the loss function to enhance the valuable information between the source and target domains. Hence, the complete loss function of the LKT module is:
\begin{equation}
\label{equa:loss}
  \mathcal{L}=\mathcal{L}_{recons}-\lambda\mathcal{L}_{mi}
\end{equation}
The process of LKT is shown in Fig.~\ref{fig:lkt}.

\myparagraph{Extension to multiple data hospitals.} 
When there are $n$ data hospitals, the \task can repeat the aforementioned Step 1 and 2 with each $\mathcal{P}_{d,i}$. Specifically, for each data hospital $\mathcal{P}_{d,i}$, \task can learn a local feature extractor $Enc_i$. Then, by aggregating $n$ local feature extractors learned from $n$ data hospitals, the local samples' final enriched representations of \task become,
\begin{equation}
  \tilde{H}_t^{nl} = \langle H_t^{nl}, Enc_1(H_t^{nl}), \cdots, Enc_n(H_t^{nl}) \rangle
\end{equation}
Our mechanism allows the \task to perform FRL and LKT with each $\mathcal{P}_{d,i}$ to obtain enhanced representations. These enhanced representations are concatenated with the $H_t^{nl}$ to obtain the final enhanced representations $\tilde{H}_t^{nl}$. However, as the number of data hospitals increases, the concatenated representations grow, leading to significant redundant knowledge. To address this challenge, we fine-tune each $Enc_i$ based on contrastive learning. For each \task-$\mathcal{P}_{d,i}$ pair, we minimize the distance between $Enc_i(H_t^{nl})$ and $Z_{t,i}^{nl}$ while maximize the distance between $Enc_i(H_t^{nl})$ and $\{Z_{t,k}^{nl};k\in n,k\ne i\}$ from other data hospitals:
\begin{equation}
    \mathcal{L}_{cl}=-\log \frac{\exp(Dist(Enc_i(H_t^{nl}),Z_{t,i}^{nl})/\tau)}{\sum_{k=1}^n\exp(Dist(Enc_k(H_t^{nl}),Z_{t,k}^{nl})/\tau)}
\end{equation}
where $\tau$ is a temperature coefficient. The goal of this approach is to enable each $Enc_i$ to learn unique representations from the view of \task and $\mathcal{P}_{d,i}$, thereby reducing redundant knowledge.

The procedure steps of \method can be found in Alg.~\ref{alg}.

\begin{algorithm}[htbp]
\caption{\method mechanism}
\label{alg}
\begin{algorithmic}[1]
\Function{FRL}{$H_t,H_d$}
\State $H_t^{ol},H_d^{ol}\gets PSI(H_t,H_d)$
\State $H_{fed}^{ol}\gets FedSVD(H_t^{ol},H_d^{ol})$
\State Return $X_{fed}^{ol}$
\EndFunction
\Statex
\Function{LKT}{$H_t,H_{d1},H_{d2},\cdots,H_{dn}$}
\State \textbf{\ding{172} Model Training}
\For{$i$ in range($n$)}
    \State Initialize an auto-encoder $Enc_i$-$Dec_i$
    \State Initialize $\Phi_i$
    \State $H_{fed,i}^{ol}\gets FRL(H_t,H_{di})$
    \While{stopping epoch not met}
        \State $Z_{t,i}^{nl}=attn(Enc_i(H_t^{nl}),H_{fed,i}^{ol}\otimes \Phi_i)$
        \State $\mathcal{L}_{recons}=||H_t^{ol}-Dec_i(Enc_i(H_t^{ol}))||_2$
        \State $\mathcal{L}_{mi} = MI(Enc_i(H_t^{nl}),Z_{t,i}^{nl})$
        \State $\mathcal{L}=\mathcal{L}_{recons}-\lambda\mathcal{L}_{mi}$
        \State $Enc_i$-$Dec_i\gets Backward(\mathcal{L})$
        \State $\Phi_i \gets Backward(\mathcal{L})$
    \EndWhile
\EndFor
\State \textbf{\ding{173} Model Fine-tuning}
\For{$i$ in range($n$)}
    \State $Z_{t,i}^{nl}=attn(Enc_i(H_t^{nl}),H_{fed,i}^{ol}\otimes \Phi_i)$
    \While{stopping epoch not met}
        \State $\mathcal{L}_{cl}=CL(Enc_i(H_t^{nl}),Z_{t,i}^{nl})$
        \State $Enc_i\gets Backward(\mathcal{L}_{cl})$
    \EndWhile
\EndFor
\State \textbf{\ding{174} Feature Augmentation}
\State $\tilde{H}_t^{nl}\gets\langle H_t^{nl}, Enc_1(H_t^{nl}), \cdots, Enc_n(H_t^{nl}) \rangle$
\State Return $\tilde{H}_t^{nl}$
\EndFunction
\end{algorithmic}
\end{algorithm}

\subsection{Security and Privacy}

Security and privacy are key factors to consider in FL mechanism design.
While \method is a knowledge transfer framework that incorporates existing VFL algorithms, the security and privacy protection levels are mainly dependent on the included VFL algorithm. In particular, the FRL module (Sec.~\ref{subsec:frl}) is the key part to determine the overall security and privacy levels of \method, as cross-party communications and computations are only conducted in this step.
Currently, we implement the FRL module with SOTA VFL representation learning methods including, FedSVD~\cite{chai2022practical} and VFedPCA~\cite{cheung2022vertical}. FedSVD uses two random orthogonal matrices to mask the original data. The third-party server can only use the masked data of each party to obtain SVD result. The third-party server of VFedPCA only needs to use the eigenvectors and eigenvalues of each party's data for weighted summation. All of these methods protect privacy by preventing direct use of data by non-holders. Due to the page limitation, readers may refer to the original papers \cite{chai2022practical,cheung2022vertical} for specific security and privacy analysis.

\subsection{Updating}
\label{subsec:updating}

In general, \method is efficient to update without the need to completely re-running three steps for all the hospitals.

\textit{\textbf{Local Incremental Learning} - New task samples}. 
\method trains a dedicated LKT for local samples from each target domain. When new samples are introduced, \task can directly apply the pre-trained LKT for efficient knowledge transfer, eliminating the need for retraining or further communication with other data hospitals.

\textit{\textbf{Task Independence} - New tasks}. Similar to new samples, if \task has a new task label to predict, \task also does not need to communicate with other hospitals. \task can simply repeat Step 3 of \method using the new task label, ensuring adaptability to evolving prediction requirements.
  
\textit{\textbf{Knowledge Extensibility} - New data hospitals}. 
\task can learn a new local feature enrichment function $Enc$ from the new \data (repeat Steps 1 and 2 with the new \data), and then enrich the representation as $\tilde{H}_t^{nl} = \langle \tilde H_t^{nl}, Enc(H_t^{nl}) \rangle $.

\section{Evaluation}
Our experiments were performed on the workstation using NVIDIA RTX 4090, Intel Xeon(R) Gold 6430, 128GB RAM, PyTorch 1.10.0, Python 3.8 and CUDA 12.0.

\subsection{Evaluation Setup}

\textbf{Datasets and Partitions}.
We evaluate \method on the seven medical datasets: 
\begin{itemize}
    \item \textit{Medical Information Mart for Intensive Care Dataset (MIMIC-\Romannum{3})}~\cite{johnson2016mimic} provides de-identified health-related data for 58,976 patients with 15 pieces of health information as features from 2001 to 2012. Length of stays is the target of prediction and varies between 1 and 4.
    \item \textit{Cardiovascular Disease Dataset (Cardio)}~\cite{cardiovascular} consists of 70,000 patient records, each containing 11 features. The labels 1 and 0 indicate the presence or absence of cardiovascular disease, respectively.
    \item \textit{Gene Expression Cancer RNA Sequence Dataset (RNA-Seq)}~\cite{weinstein2013cancer} contains gene expression data from patients with various tumor types. It has a dimension of 801$\times$20,531, with five tumor types as prediction targets. Each sample's features represent RNA-Seq gene expression levels.
    \item \textit{Heart Attack Risk Prediction Dataset (Heart)}~\cite{heart-attack} is synthetic and includes 26 features associated with heart health and lifestyle choices for 8,763 patients. Labels 1 and 0 indicate whether a patient is at risk of a heart attack or not, respectively.
    \item \textit{Prediction of Sepsis Dataset (Sepsis)}~\cite{sepsis} includes 17 clinical and laboratory features from 45,852 patients. The task is to predict sepsis six hours before its clinical onset. A label of 1 indicates an impending sepsis event, while 0 signifies no risk.
    \item \textit{Leukemia Classification (Leukemia) dataset}~\cite{leukemia} contains 15,135 images from 118 patients with two label: 0 for normal cell and 1 for leukemia cell.
    \item \textit{Chest X-Ray Images (Pneumonia) dataset}~\cite{pneumonia} contains 5,863 X-ray images and 2 classes (pneumonia/normal).
\end{itemize}

The used datasets have been widely used in previous FL-empowered healthcare studies~\cite{HuangW023,QayyumAAA022,YangZHSC21}.  
The dataset summary and detailed data partitions is provided in Appendix.
We assume a scenario with one \task and one \data by default, which aligns with the standard setup in VFL~\cite{PangYSW23,LiuKZPHYOZY24}. 
The five datasets--MIMIC-\Romannum{3}, Cardio, RNA-Seq, Heart, and Sepsis--are utilized to evaluate the effectiveness of intra-domain knowledge transfer. The two datasets, Leukemia and Pneumonia, are used to assess the performance of cross-domain knowledge transfer. 
In the intra-domain setting, both $H_t^{nl}$ and $H_t^{ol}$ are derived from the same domain, share an identical feature space, and are randomly sampled from the original dataset. 
Following the existing cross-domain settings~\cite{ZhaoZXZGZ23,ChenDCQH20}, we define the target domain data, $H_t^{nl}$, as cropped images focusing on the central area, while the source domain data, $H_t^{ol}$, consists of complete images. During data partition, the target domain is assigned two types of labels, whereas the source domain retains a single type of label.

\textbf{Baselines}.
To verify the effectiveness of our mechanism, we compare with six baselines:

\begin{itemize}
  \item \textit{Local}: This baseline leverages only the \task's local non-overlapping $H_t^{nl}$ for training the task-specific prediction.
  \item \textit{FTL~\cite{Liu2020ASF}}: FTL is an end-to-end FL method for transferring knowledge to local samples. Specifically, based on shared samples, FTL maps different parties' raw features to a common feature space to achieve knowledge transfer.
  \item \textit{VFL-Infer~\cite{RenYC22}}: VFL-Infer first learns a federated model on shared samples and then learns a local model (for local samples) by considering both ground-truth labels and soft labels produced by the federated model.
  \item \textit{FedCVT~\cite{KangLL22}}: FedCVT leverages existing representations to estimate missing features for \task and generates pseudo-labels for unlabeled samples, thereby creating an expanded training dataset.
  \item \textit{FedHSSL~\cite{fedhssl}}: FedHSSL first uses cross-party SSL to train a cross-party encoder on $H_t^{ol}$ and $H_d^{ol}$, and then uses the cross-party encoder to train a local encoder on $H_t^{ol}$ and $H_t^{nl}$ to perform augmentation.
  \item \textit{VFedTrans~\cite{HuangW023}}: VFedTrans~\cite{HuangW023} facilitates knowledge transfer in a task-independent manner through local representation distillation.
\end{itemize}
The default settings for these baselines follow the configuration used in the original paper. While FTL, VFL-Infer, FedCVT, FedHSSL, and VFedTrans can be used to assist local non-overlapping samples' learning in VFL, they do not address two key challenges: feature heterogeneity (FH) and label heterogeneity (LH). 

\textbf{Training Setup}.
\label{subsec:training}
\method's training involves three main modules: the FRL module, the LKT module, and task-specific models.

\textit{FRL modules}. We use two VFL techniques, FedSVD and VFedPCA, FedSVD and VFedPCA, with FedSVD as the default method. We set block\_size to 100 in FedSVD. For VFedPCA, we configure iter\_num to 100, period\_num to 10, and warm\_start to True.

\textit{LKT modules}. 
For the LKT modules, we use the Adam optimizer~\cite{KingmaB14} with the following settings: learning\_rate = 0.001, batch\_size = 100, and 30 epochs. We use $\beta_0=1e-3$ and $\beta_1=1e-1$ to balance $\mathcal{L}_{recons}$ and $\mathcal{L}_{mi}$. 
The default LKT method is the vanilla Auto-Encoder (AE). Additionally, we conduct experiments with alternative LKT methods, including Variational Auto-Encoder (VAE)~\cite{kingma2013auto} and Wasserstein Auto-Encoder (WAE)~\cite{TolstikhinBGS18},  to assess the robustness of our framework. The key parameters for these three modules are summarized in Appendix.

\textit{Task-specific models.}
For all the datasets, when training the task-specific model, we choose $80\%$ of $H_t^{nl}$ as the training set and $20\%$ as the test set. 
In order to prevent the interference of random seeds, we carry out experiments under $10$ different random seeds and compute the average prediction results. 

Note that traditional ML models often perform efficiently and effectively in many medical tasks~\cite{morrill2019signature}. 
In our experiments, we use the XGBoost~\cite{Chen2016XGBoostAS} and Convolutional Neural Networks (CNN)~\cite{OSheaN15} as the default algorithm. We also test the other popular algorithms, including Adaboost~\cite{freund1999short}, multi-Layer perceptron (MLP), TabNet~\cite{ArikP21} and VGG16~\cite{SimonyanZ14a} for robustness checks in Sec.~\ref{subsec:robustness}. 
For XGBoost and Adaboost, we use grid search to find the optimal parameters.
The parameters of the rest downstream model are summarized in Appendix.

We evaluate \method by addressing the following key research questions:
\begin{itemize}
    \item \textbf{RQ1.} How effective is \method's intra-domain knowledge transfer?
    \item \textbf{RQ2.} How effective is \method's cross-domain knowledge transfer?
    \item \textbf{RQ3.} How does \method perform in a few-shot scenario with limited training data?
    \item \textbf{RQ4.} Can \method achieve strong performance on different FRL, LKT modules and downstream ML methods?
    \item \textbf{RQ5.} How does \method's prediction performance for non-overlapping patients and system efficiency change as the number of data hospitals increases?
\end{itemize}

\subsection{Intra-domain Main Evaluations}
\label{subsec:intro_domain_eval}

\begin{figure*}[htbp]
\centering
\begin{minipage}{1.0\textwidth}
    \begin{minipage}{0.17\textwidth}
        \centering
        \includegraphics[width=\linewidth]{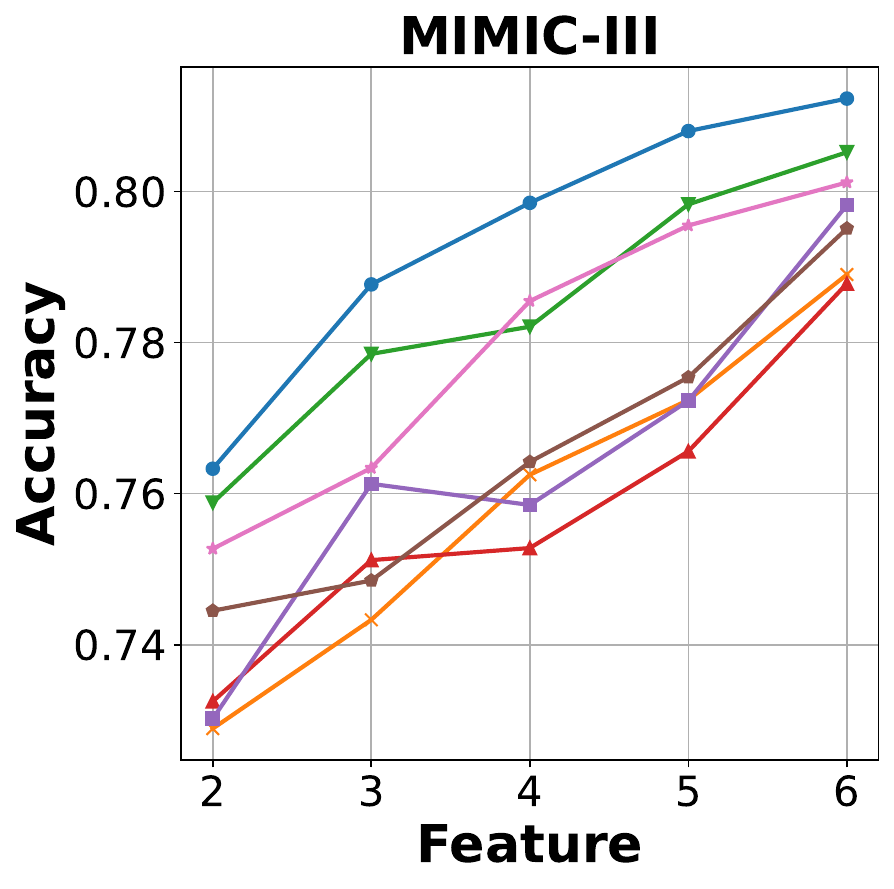}
    \end{minipage} \hfill
    \begin{minipage}{0.17\textwidth}
        \centering
        \includegraphics[width=\linewidth]{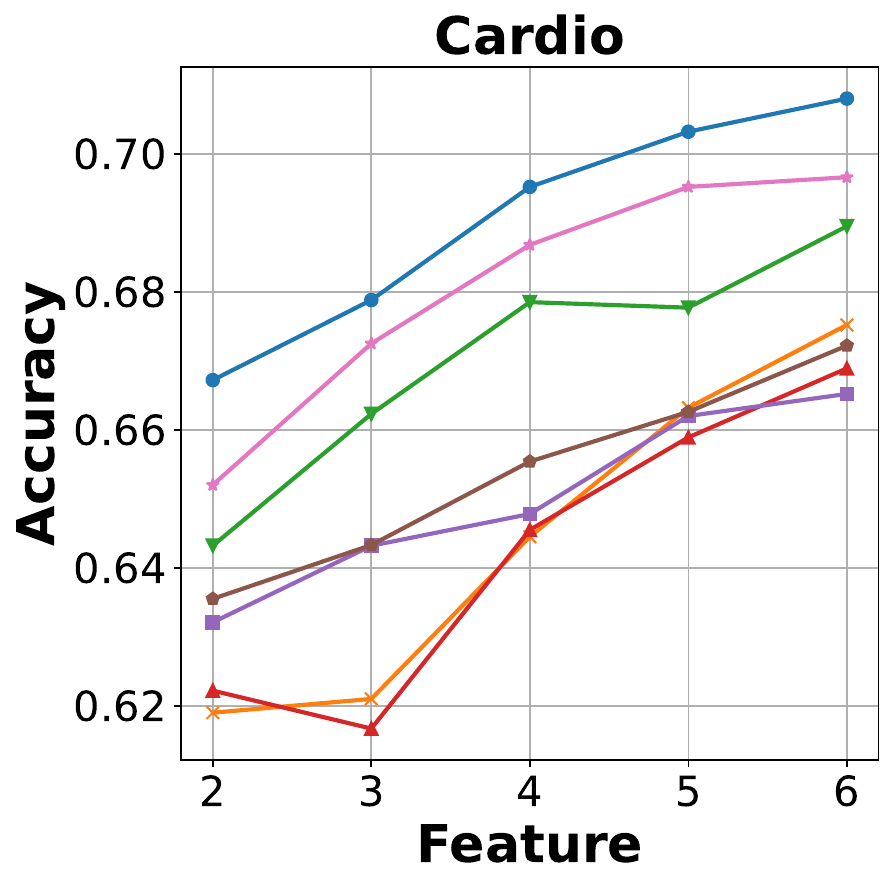}
    \end{minipage} \hfill
    \begin{minipage}{0.17\textwidth}
        \centering
        \includegraphics[width=\linewidth]{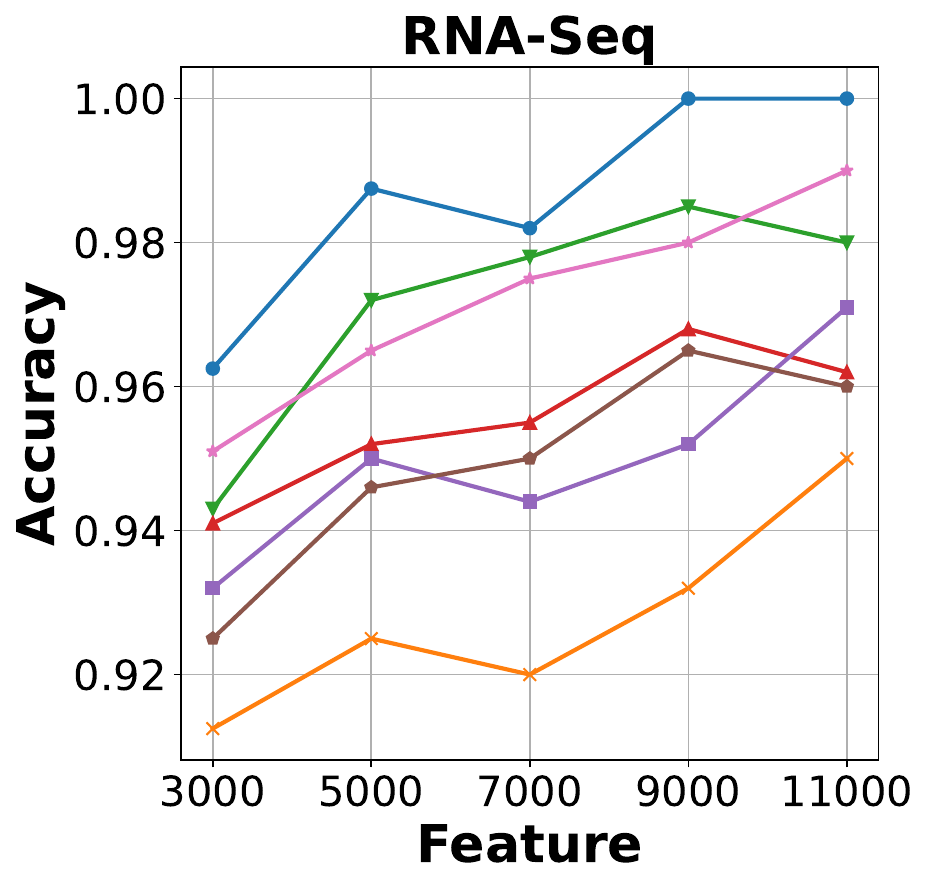}
    \end{minipage} \hfill
    \begin{minipage}{0.17\textwidth}
        \centering
        \includegraphics[width=\linewidth]{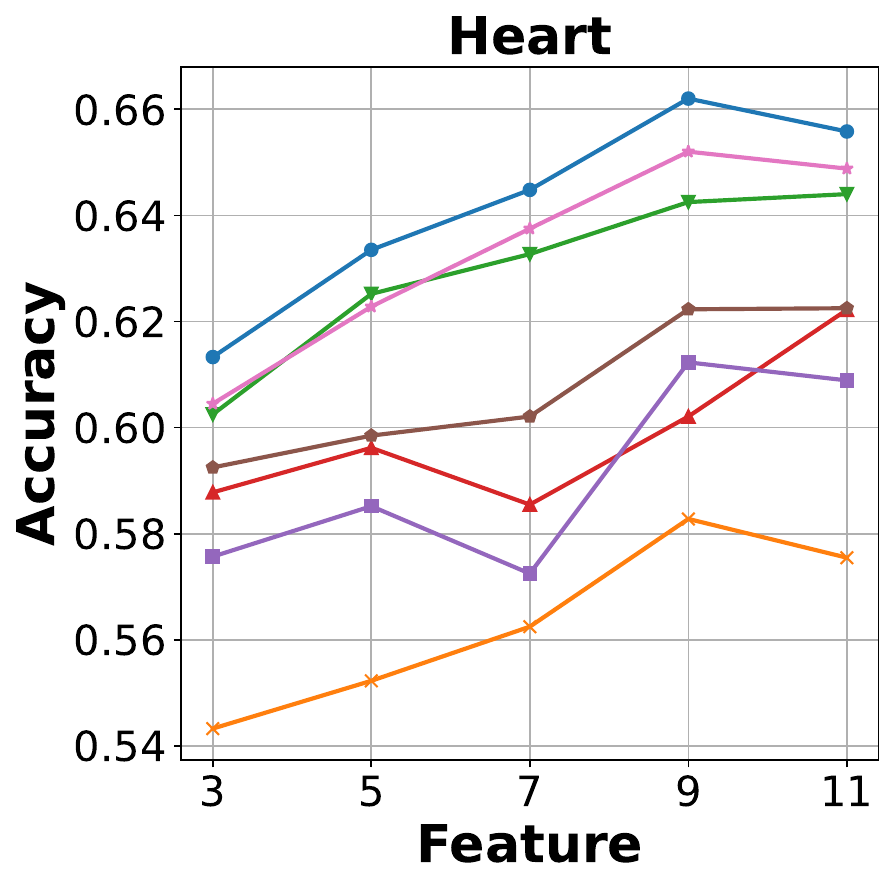}
    \end{minipage} \hfill
    \begin{minipage}{0.26\textwidth}
        \centering
        \includegraphics[width=\linewidth]{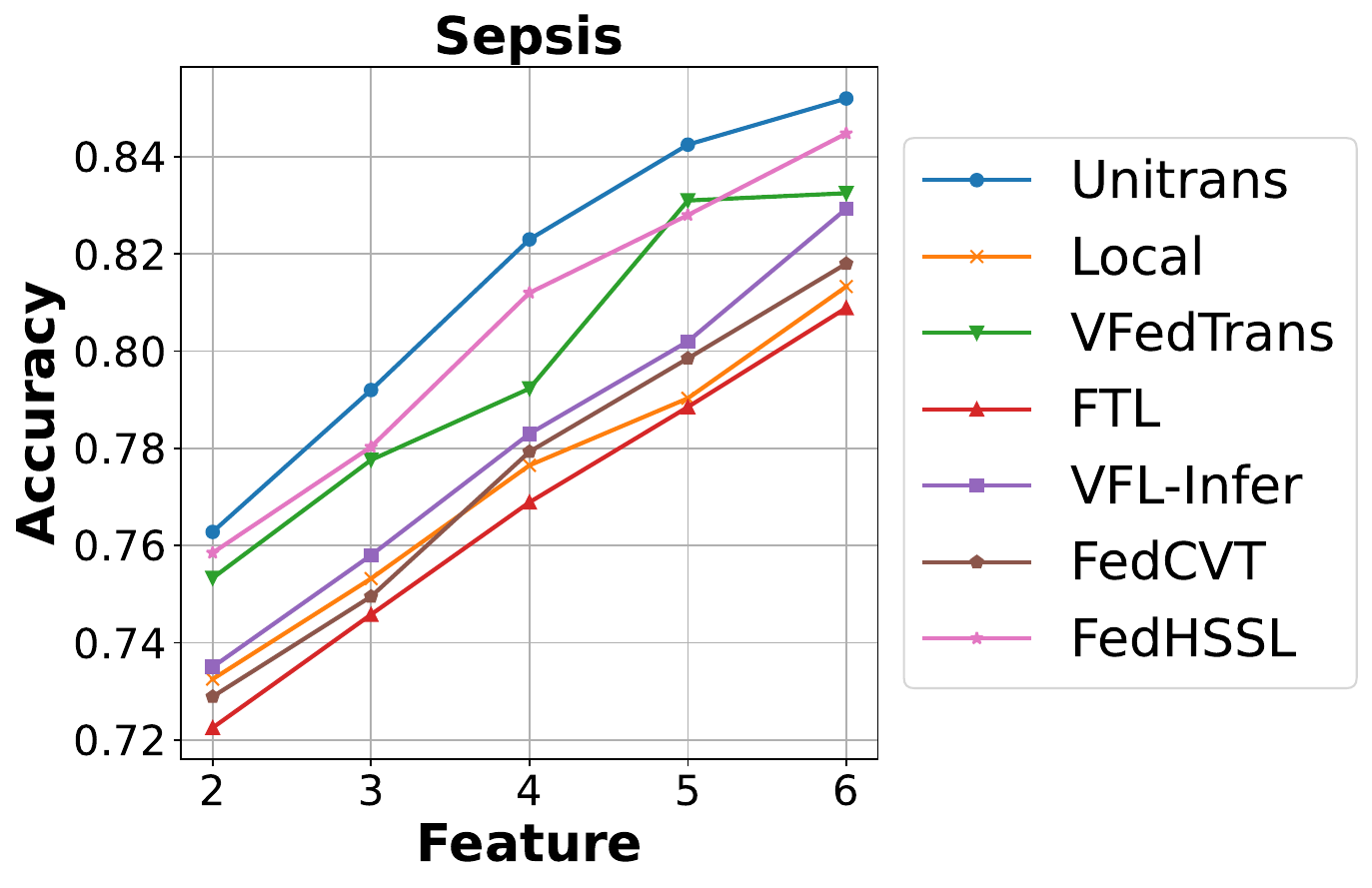}
    \end{minipage}

    \caption{Prediction accuracy by varying the feature numbers of \task.}
    \label{fig:id_vary_task_feature}
\end{minipage}
\vskip 0.25cm
\begin{minipage}{1.0\textwidth}
    \begin{minipage}{0.17\textwidth}
        \centering
        \includegraphics[width=\linewidth]{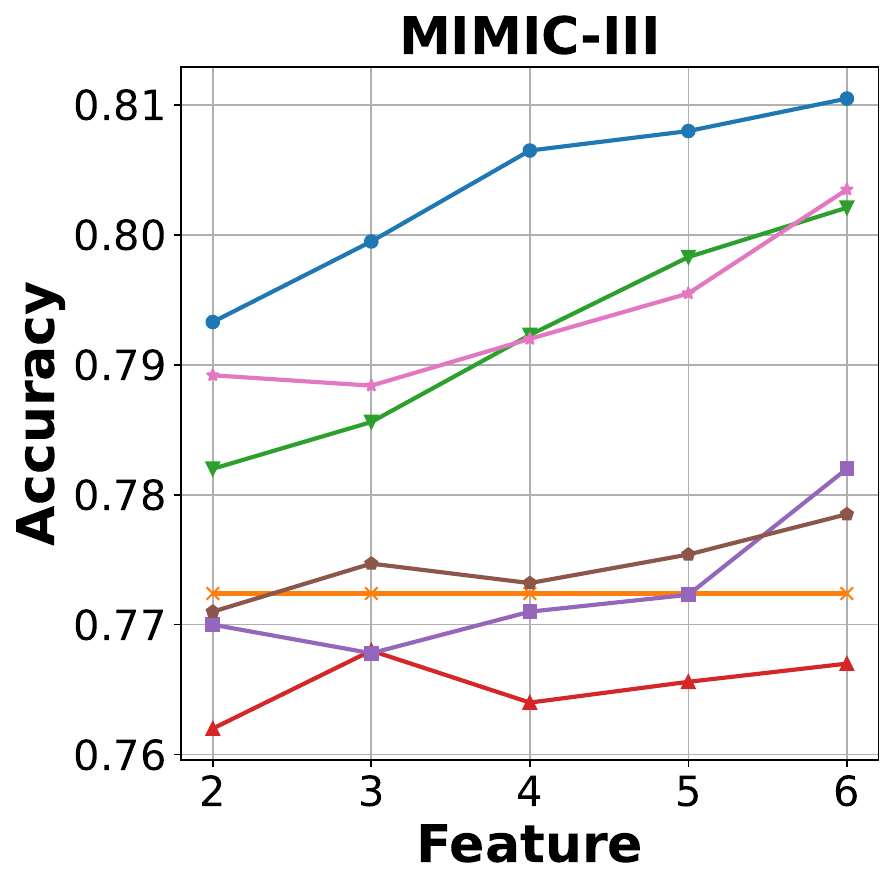}
    \end{minipage} \hfill
    \begin{minipage}{0.17\textwidth}
        \centering
        \includegraphics[width=\linewidth]{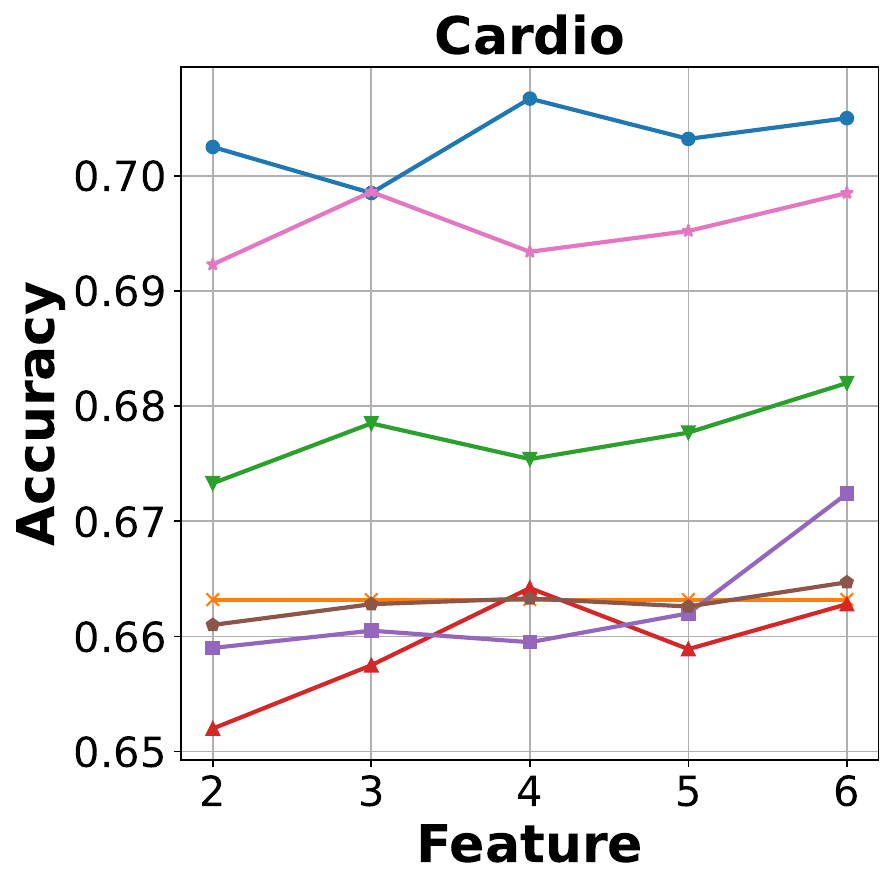}
    \end{minipage} \hfill
    \begin{minipage}{0.17\textwidth}
        \centering
        \includegraphics[width=\linewidth]{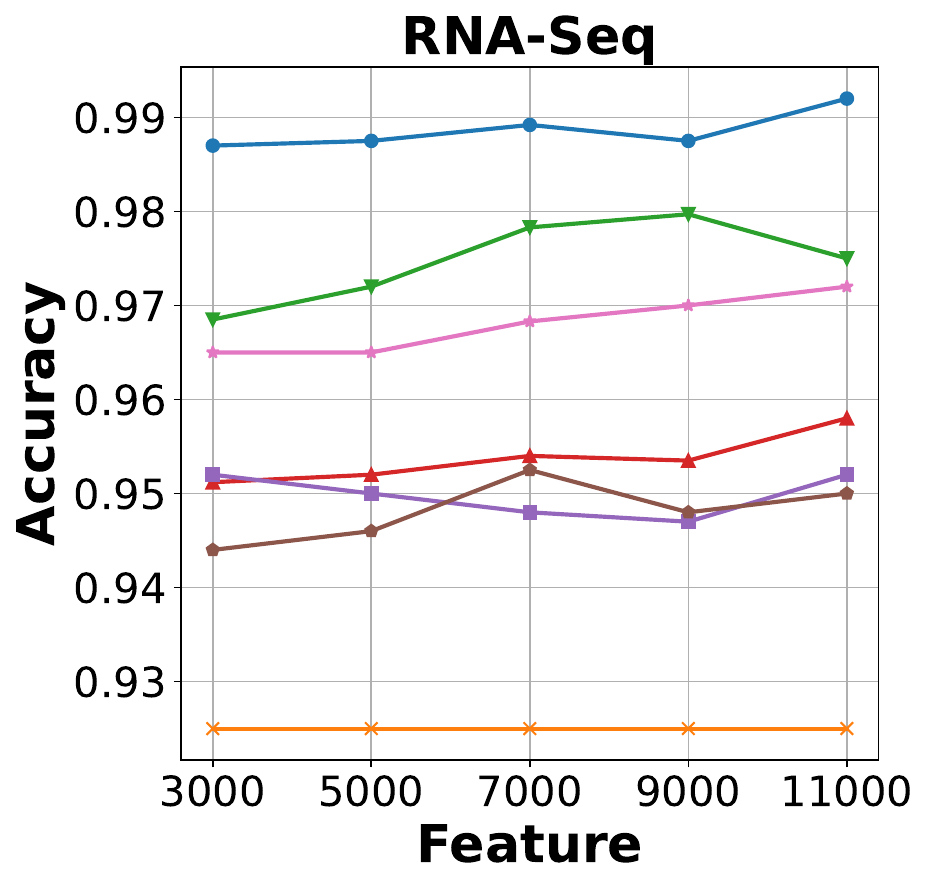}
    \end{minipage} \hfill
    \begin{minipage}{0.17\textwidth}
        \centering
        \includegraphics[width=\linewidth]{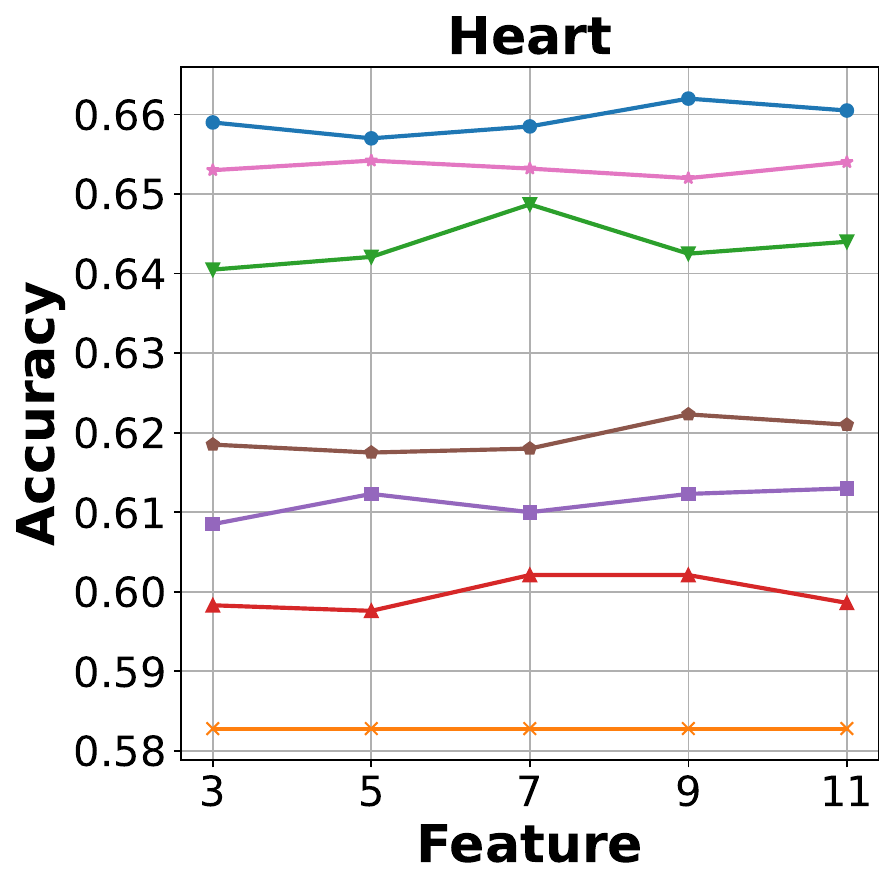}
    \end{minipage} \hfill
    \begin{minipage}{0.26\textwidth}
        \centering
        \includegraphics[width=\linewidth]{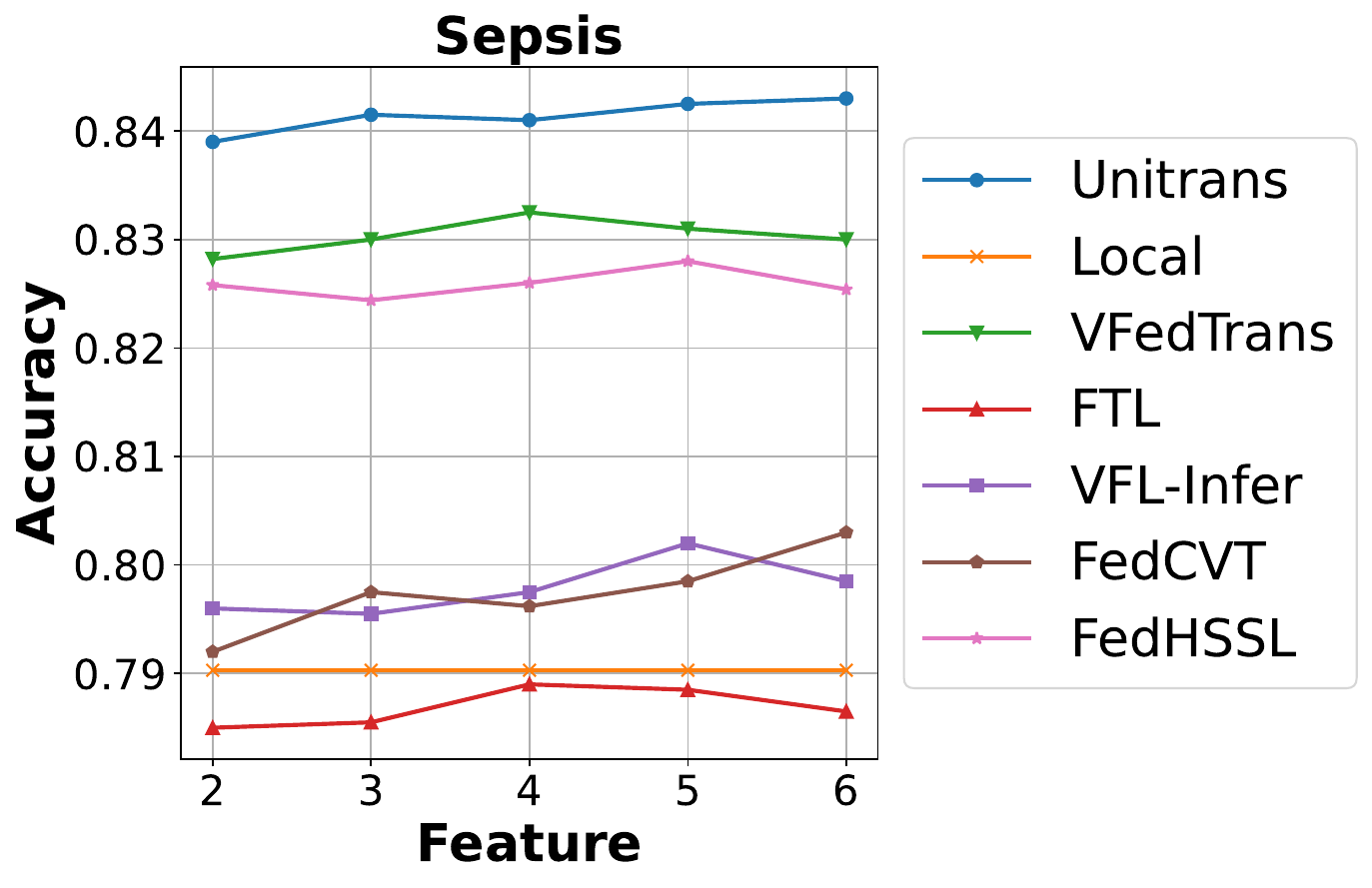}
    \end{minipage}

    \caption{Prediction accuracy by varying the feature numbers of \data.}
    \label{fig:id_vary_data_feature}
\end{minipage}
\vskip 0.25cm
\begin{minipage}{1.0\textwidth}
    \begin{minipage}{0.17\textwidth}
        \centering
        \includegraphics[width=\linewidth]{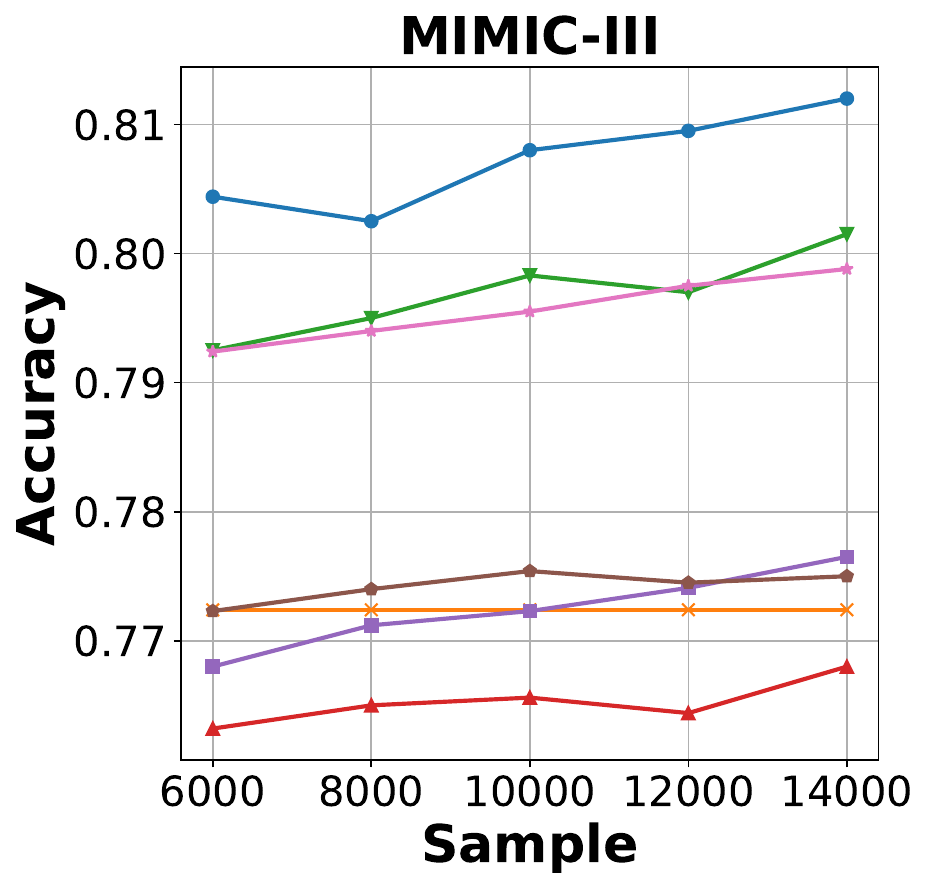}
    \end{minipage} \hfill
    \begin{minipage}{0.17\textwidth}
        \centering
        \includegraphics[width=\linewidth]{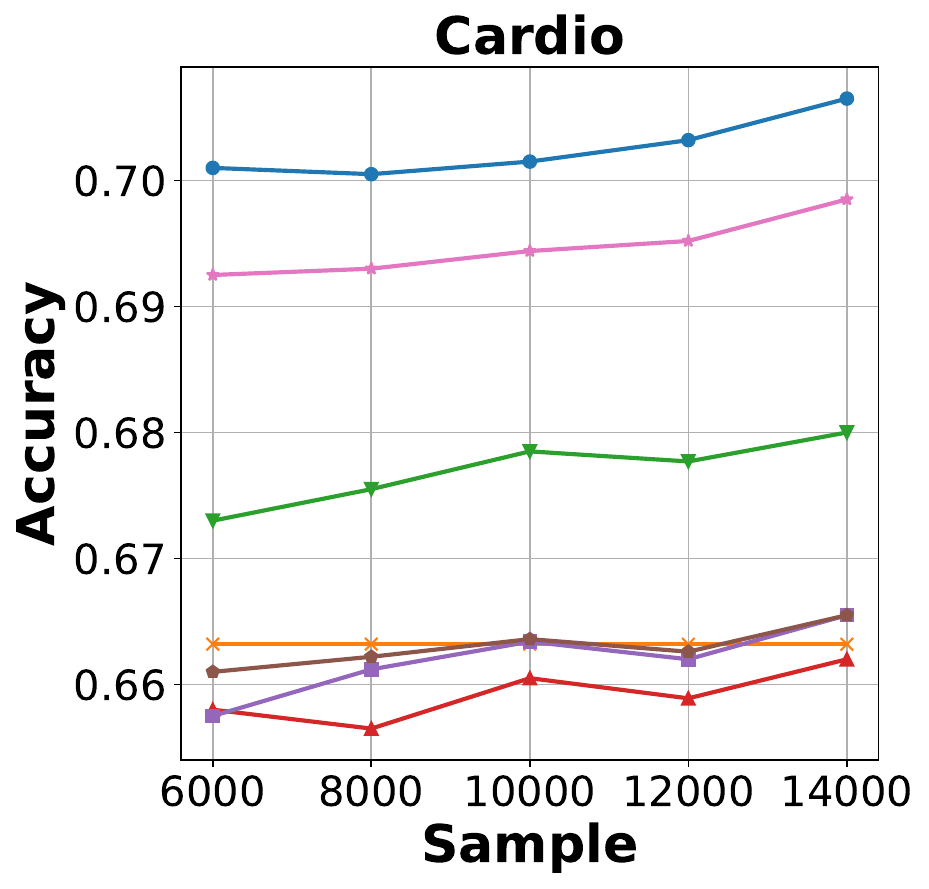}
    \end{minipage} \hfill
    \begin{minipage}{0.17\textwidth}
        \centering
        \includegraphics[width=\linewidth]{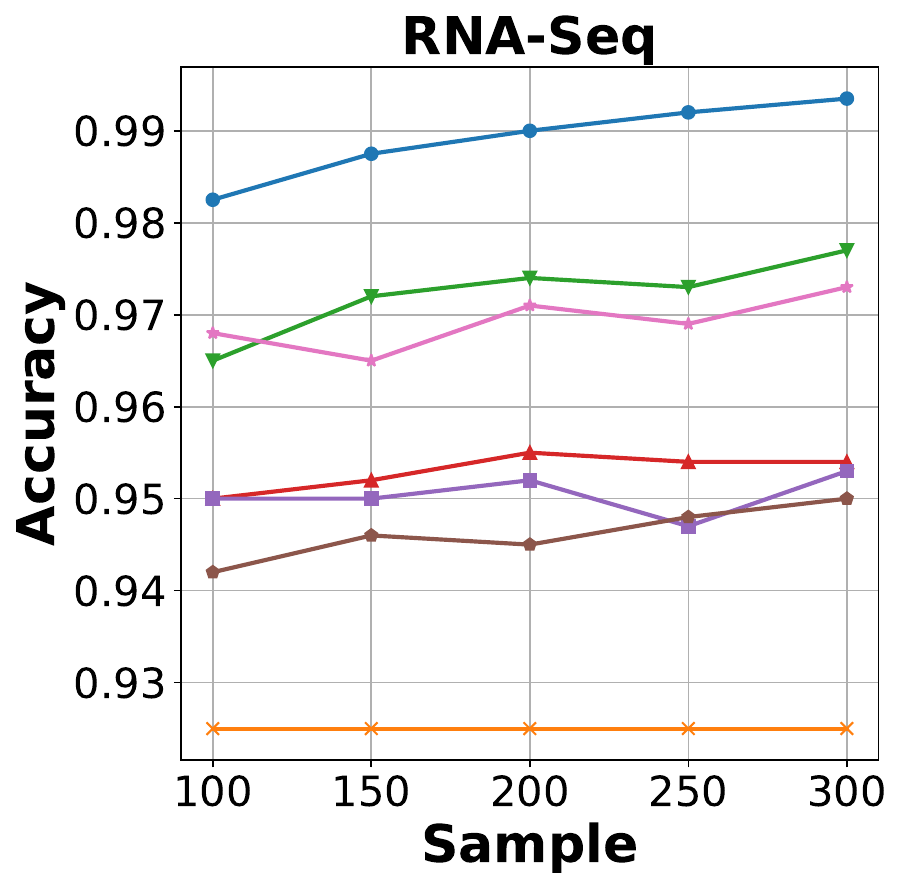}
    \end{minipage} \hfill
    \begin{minipage}{0.17\textwidth}
        \centering
        \includegraphics[width=\linewidth]{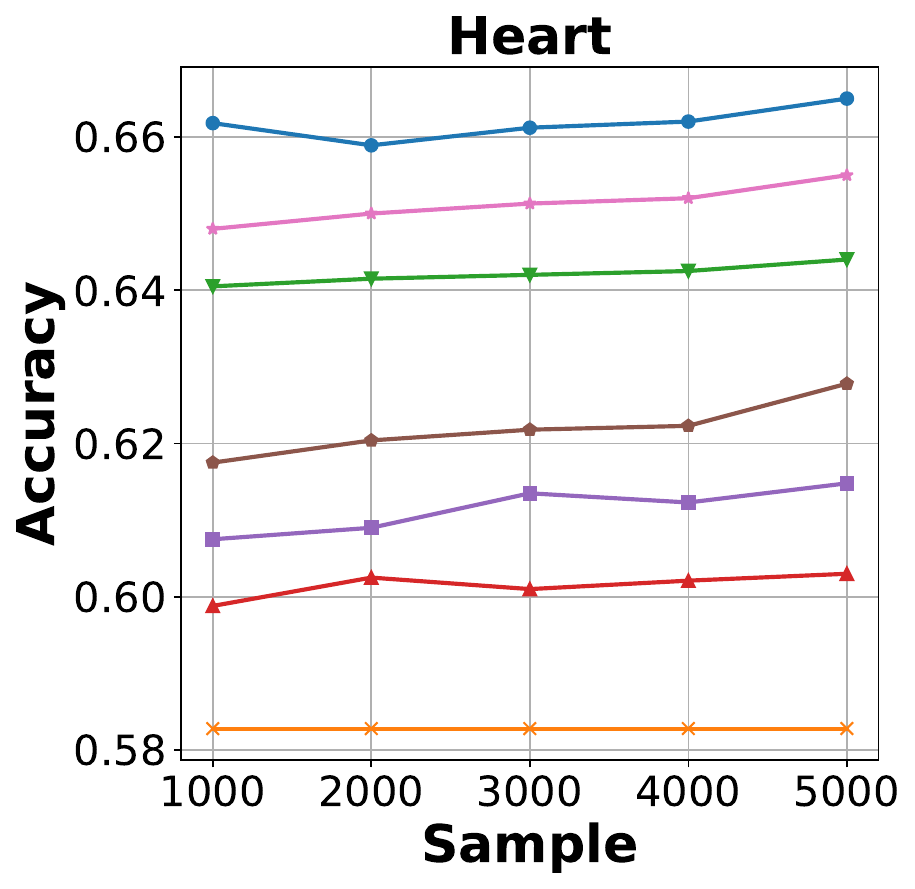}
    \end{minipage} \hfill
    \begin{minipage}{0.26\textwidth}
        \centering
        \includegraphics[width=\linewidth]{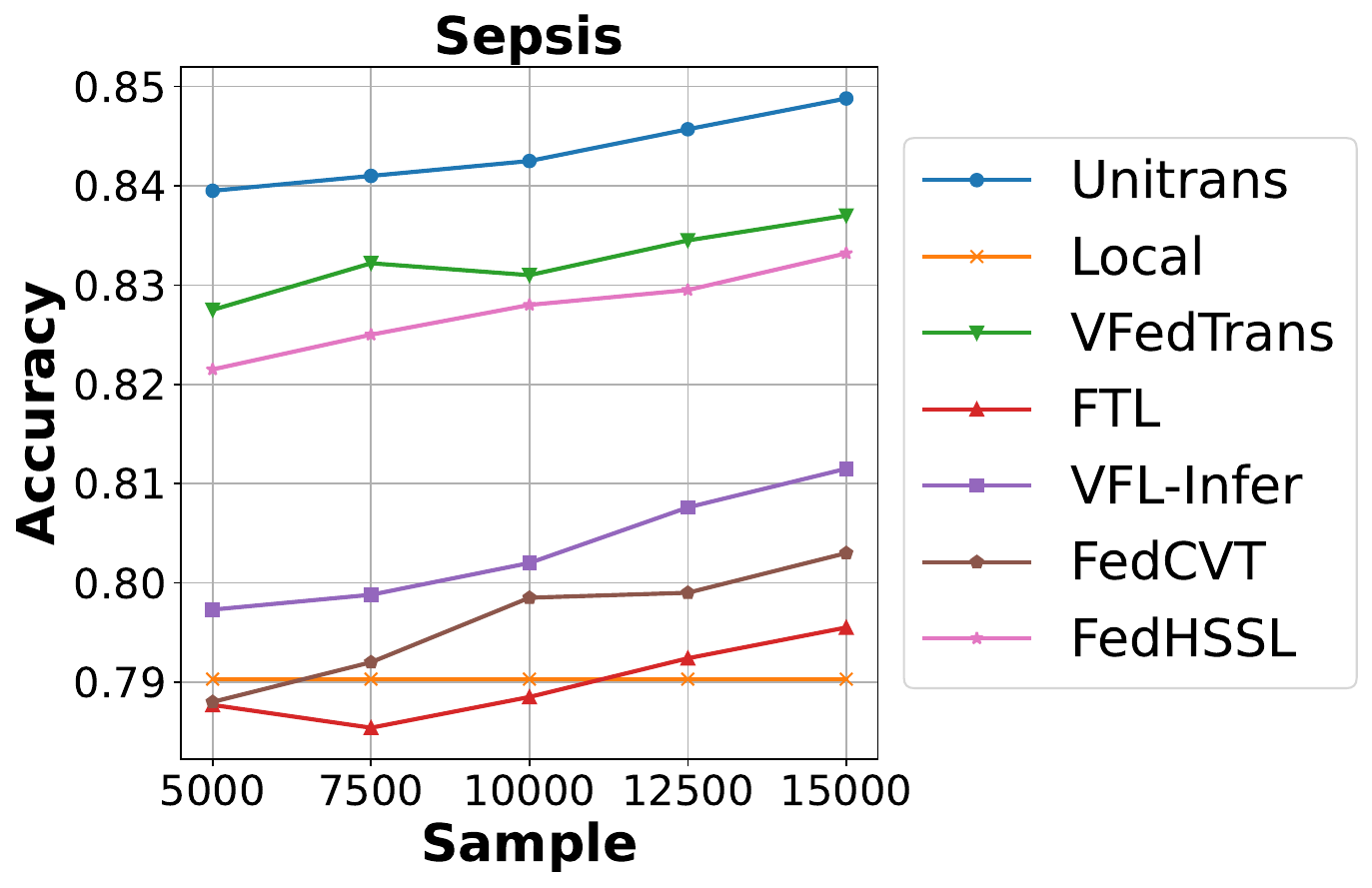}
    \end{minipage}

    \caption{Prediction accuracy by varying the number of overlapping samples between \task and \data.}
    \label{fig:id_vary_ol_sample}
\end{minipage}
\end{figure*}

We first verify the effectiveness of \method in intra-domain knowledge transfer.
Fig.~\ref{fig:id_vary_task_feature} depicts the prediction performance on five datasets by varying the number of features in \task. Intra-domain learning requires that the input dimensions of the data are consistent, so the feature dimensions of $H_t^{ol}$ and $H_t^{nl}$ will change synchronously. 
The results show that as the feature scale of \task increases, the prediction accuracy of all methods improves. Due to more effective intra-domain knowledge transfer, \method achieves the highest accuracy. It is worth noting that when the number of features is relatively low, existing baselines struggle to make effective predictions. However, the feature augmentation strategy provided by \method enhances prediction performance and is particularly beneficial for hospitals with limited medical resources.

Fig.~\ref{fig:id_vary_data_feature} illustrates the prediction performance across five datasets as the number of features in \data is varied. The accuracy of \method consistently improves as the number of features in \data increases. Notably, \method achieves higher classification accuracy compared to the baselines, suggesting that it can transfer knowledge from the rich feature set of \data more efficiently.

Fig.~\ref{fig:id_vary_ol_sample} shows how our mechanism performs by changing the number of overlapping samples between \task and \data. 
We observe that the performance gets better as there are more shared samples.
This also validates the effectiveness of \method: with more knowledge sources (i.e., shared samples), our transfer can always be better.

\begin{tcolorbox}[left=0mm, right=0mm, top=0mm, bottom=0mm]
\textbf{Answer to \textbf{RQ1}: \method effectively transfers cross-party knowledge to non-overlapping samples within the domain. As the knowledge base grows, \method improves its ability to make more accurate predictions for downstream tasks.}
\end{tcolorbox}

\subsection{Cross-domain Main Evaluations}

Then we verify the effectiveness of \method in cross-domain knowledge transfer. As discussed in Sec.~\ref{sec:intro}, cross-domain scenarios present significant challenges, including feature heterogeneity and label heterogeneity. To thoroughly assess \method, we separately validate its performance under each of these heterogeneous scenarios.

\begin{figure*}[ht]
\centering
\begin{minipage}{1.0\textwidth}
    \begin{minipage}{0.48\textwidth}
        \begin{minipage}{0.48\linewidth}
            \centering
            \includegraphics[width=1\linewidth]{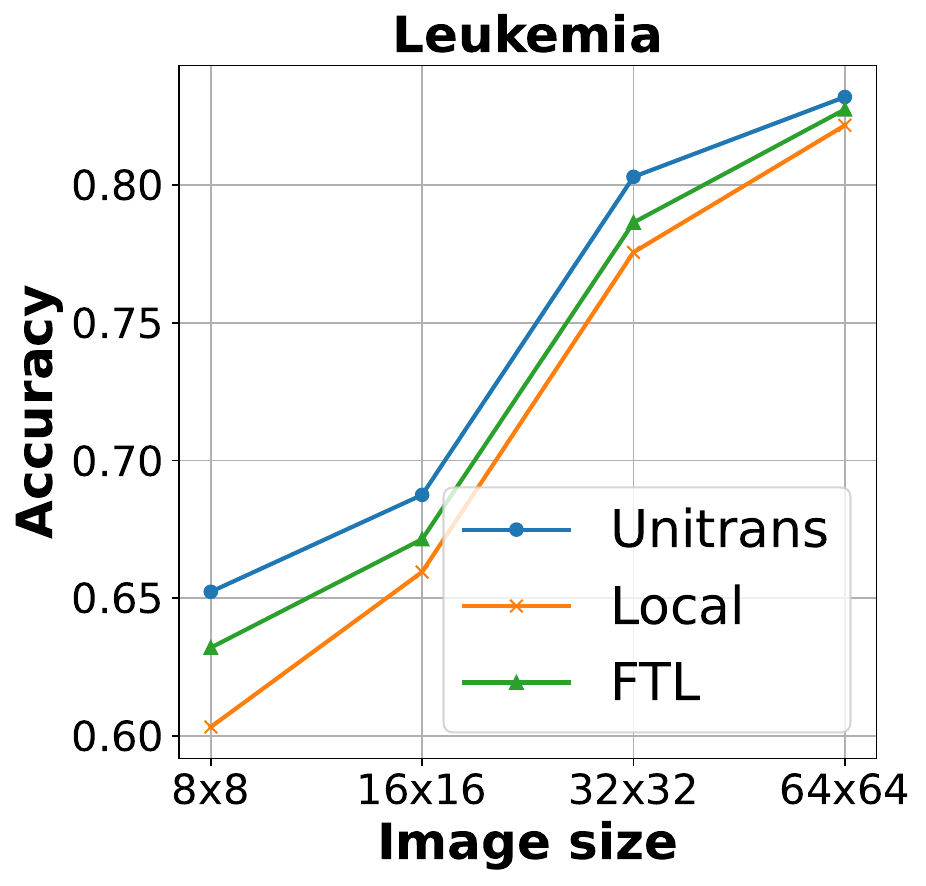}
        \end{minipage}
        \begin{minipage}{0.48\linewidth}
            \centering
            \includegraphics[width=1\linewidth]{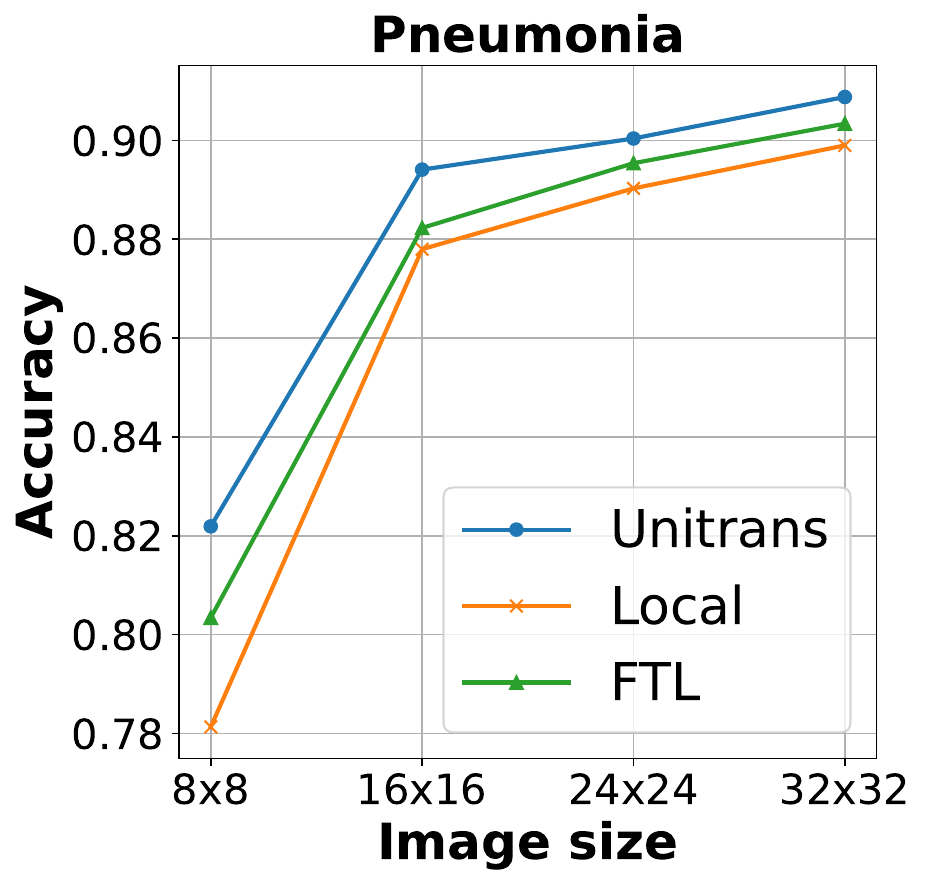}
        \end{minipage}\hfill
        \caption{Prediction accuracy of the target domain at different image sizes.}
        \label{fig:cross_feature_imgsize}
    \end{minipage}
    \begin{minipage}{0.48\textwidth}
        \begin{minipage}{0.48\linewidth}
            \centering
            \includegraphics[width=1\linewidth]{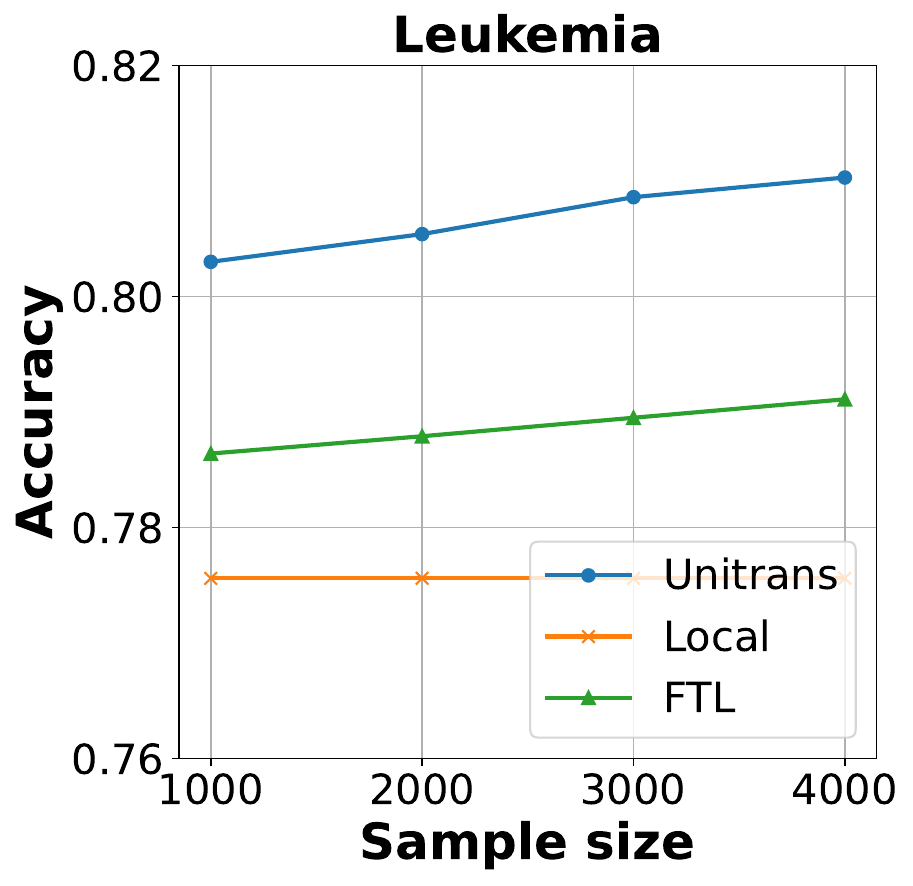}
        \end{minipage}
        \begin{minipage}{0.48\linewidth}
            \centering
            \includegraphics[width=1\linewidth]{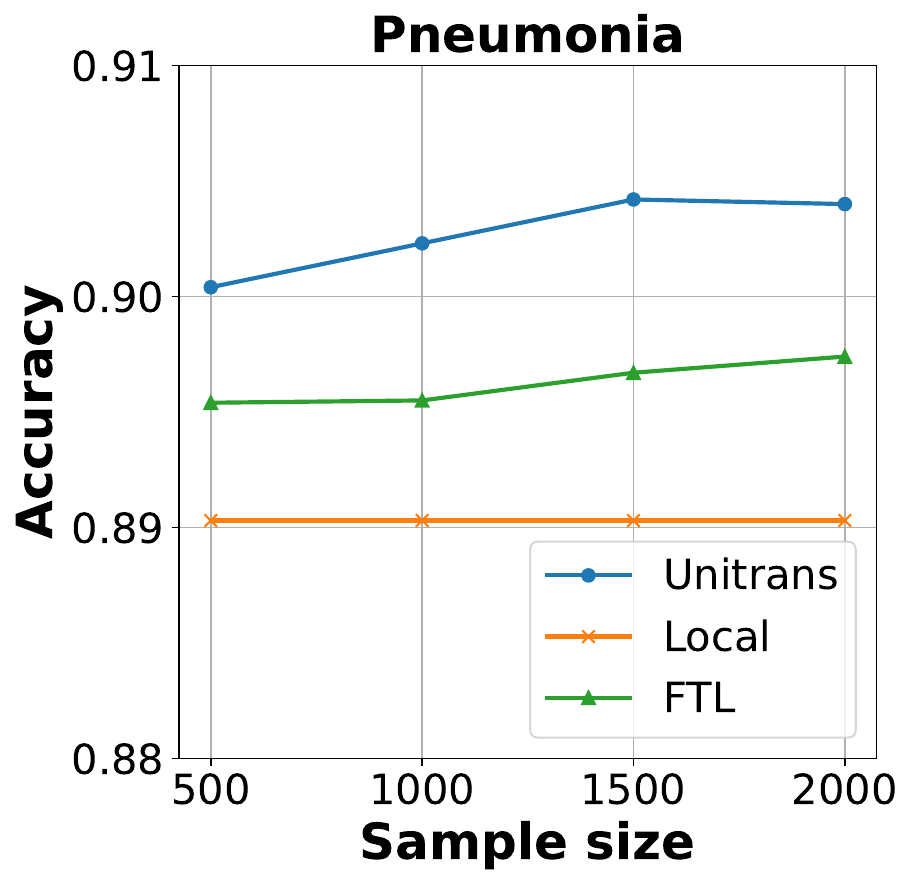}
        \end{minipage}\hfill
        \caption{Prediction accuracy of the target domain at different sample sizes.}
        \label{fig:cross_feature_sample_size}
    \end{minipage}
\end{minipage}
\begin{minipage}{1.0\textwidth}
    \begin{minipage}{0.47\textwidth}
        \vspace{10pt}
        \begin{minipage}{0.48\linewidth}
            \centering
            \includegraphics[width=1\linewidth]{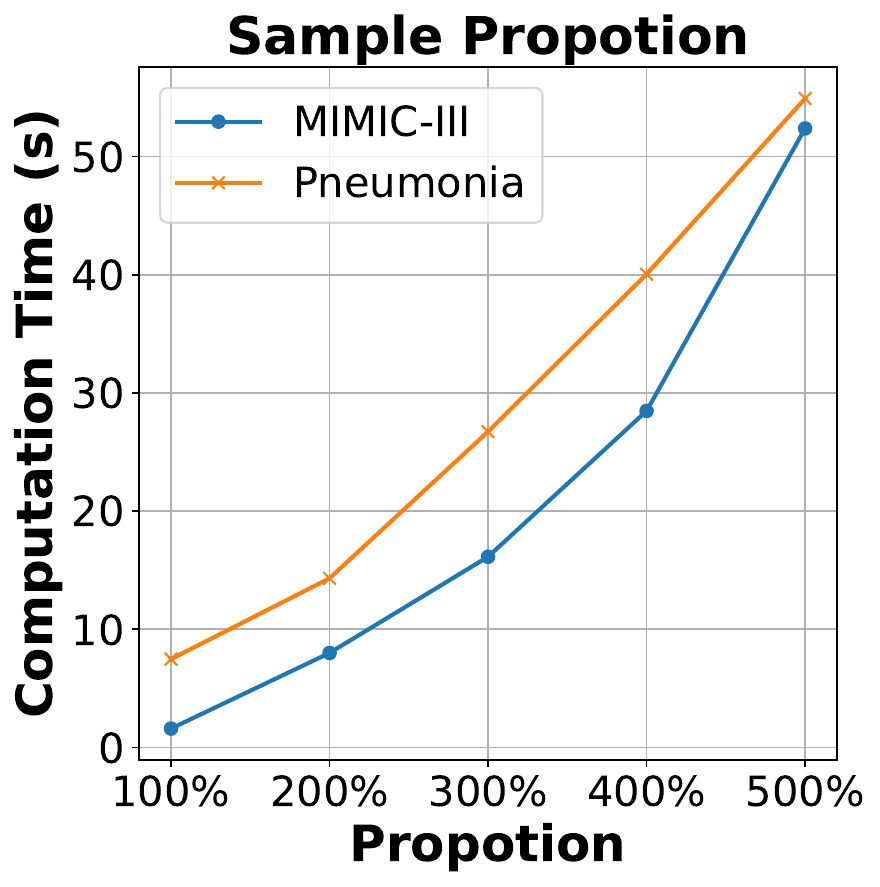}
        \end{minipage}
        \begin{minipage}{0.48\linewidth}
            \centering
            \includegraphics[width=1\linewidth]{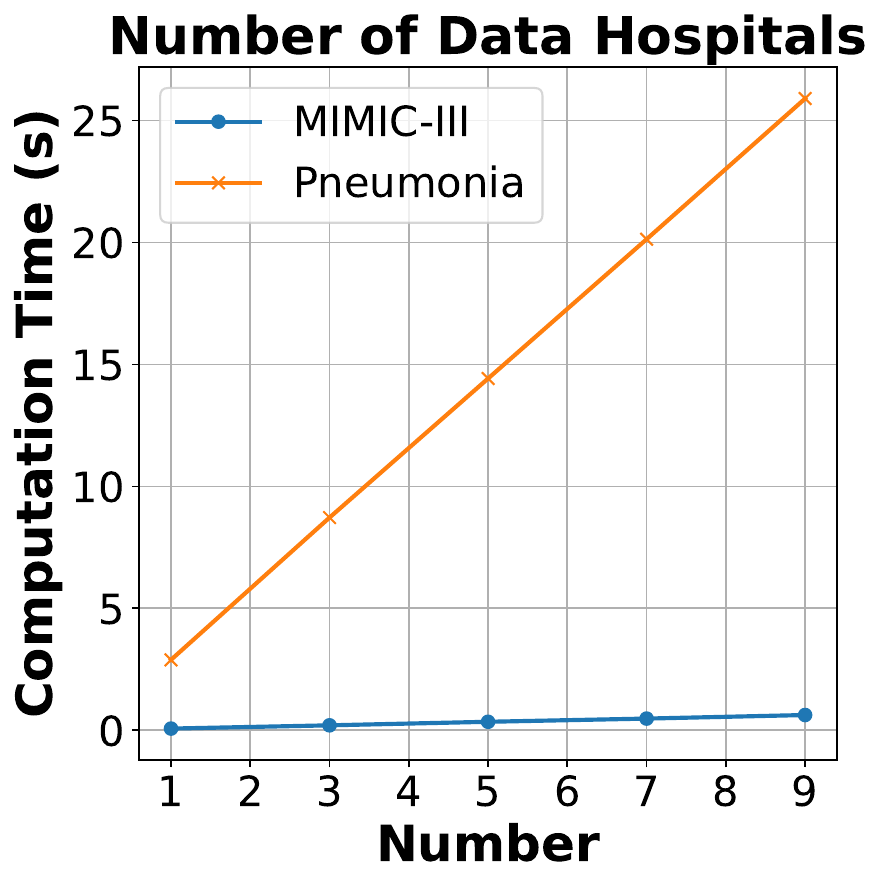}
        \end{minipage}\hfill
        \vspace{-2pt}
        \caption{Computation time by varying the number of $\mathcal{P}_d$s and the number of overlapping samples.}
        \label{fig:scala_time}
    \end{minipage}
    \begin{minipage}{0.48\textwidth}
        \begin{minipage}{0.48\linewidth}
            \centering
            \includegraphics[width=1\linewidth]{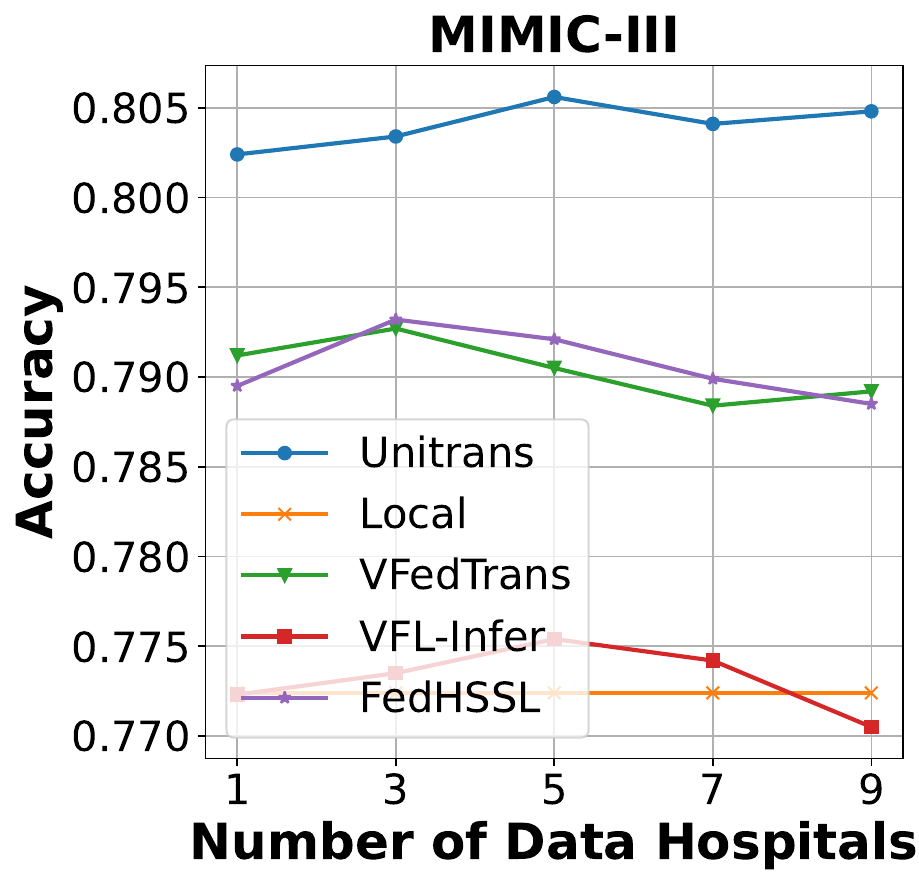}
        \end{minipage}
        \begin{minipage}{0.48\linewidth}
            \centering
            \includegraphics[width=1\linewidth]{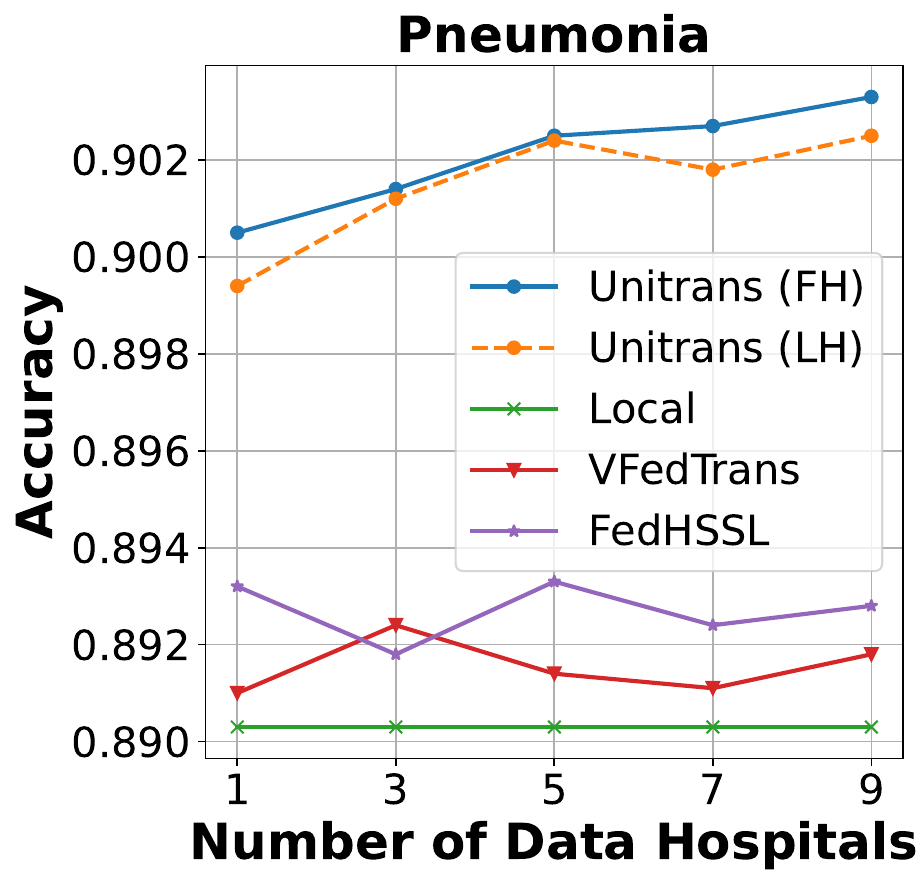}
        \end{minipage}\hfill
        \caption{Prediction accuracy by increasing the number of $\mathcal{P}_d$s.}
        \label{fig:scala_data_hospital_accuracy}
    \end{minipage}
\end{minipage}
\begin{minipage}{1.0\textwidth}
    \begin{minipage}{0.48\linewidth}
        \begin{minipage}{0.48\linewidth}
            \centering
            \includegraphics[width=1\linewidth]{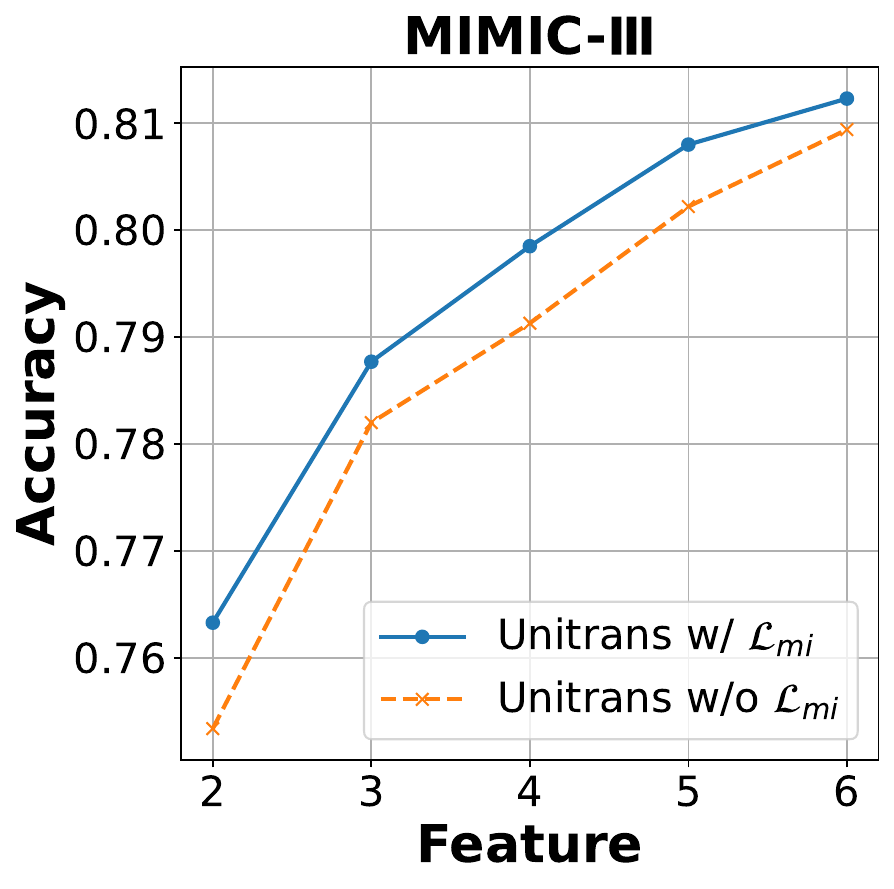}
        \end{minipage}
        \begin{minipage}{0.48\linewidth}
            \centering
            \includegraphics[width=1\linewidth]{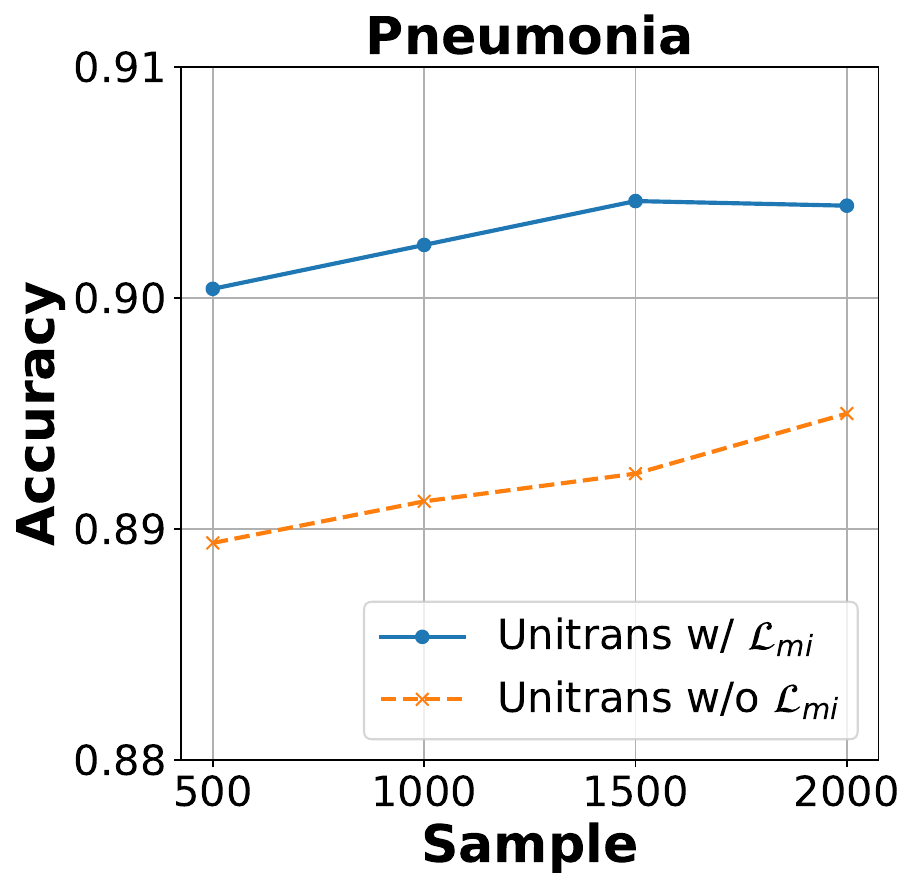}
        \end{minipage}\hfill
        \caption{Prediction accuracy of MIMIC-\Romannum{3} and Pneumonia with or without $\mathcal{L}_{mi}$.}
        \label{fig:ablation_mi}
    \end{minipage}
    \begin{minipage}{0.48\linewidth}
        \begin{minipage}{0.48\linewidth}
            \centering
            \includegraphics[width=1\linewidth]{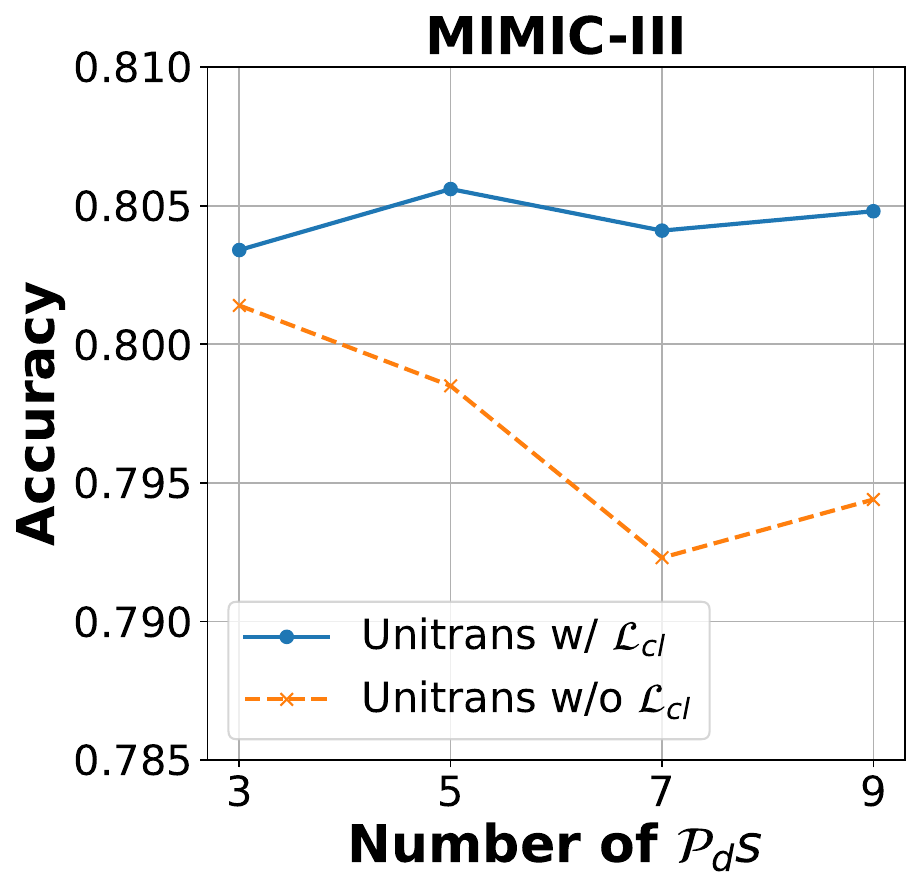}
        \end{minipage}\hfill
        \begin{minipage}{0.48\linewidth}
            \centering
            \includegraphics[width=1\linewidth]{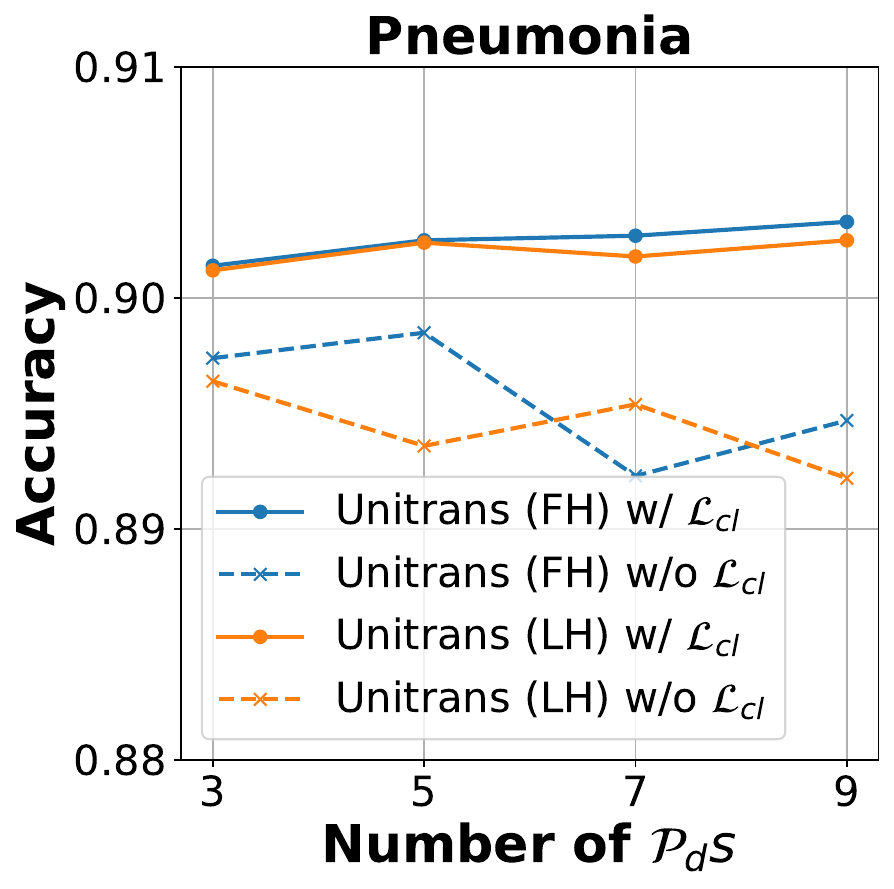}
        \end{minipage}\hfill
        \caption{Prediction accuracy by varying number of $\mathcal{P}_d$s with or without $\mathcal{L}_{cl}$.}
        \label{fig:ablation_cl}
    \end{minipage}
\end{minipage}
\end{figure*}

\textbf{Feature Heterogeneity}. As shown in Tbl.~\ref{tab:compare}, among the existing methods, only FTL is capable of addressing feature heterogeneity. To evaluate the effectiveness of \method in this context, we compare its downstream task prediction performance against Local and FTL approaches on the Leukemia and Pneumonia datasets. 
Fig.~\ref{fig:cross_feature_imgsize} illustrates the prediction performance across various target domain image sizes. The results demonstrate that \method consistently achieves the highest prediction accuracy. Notably, when the target domain image size is small (indicating insufficient knowledge), \method significantly enhances the target domain's prediction performance. This highlights \method's robustness and effectiveness in scenarios with limited knowledge. 
Fig.~\ref{fig:cross_feature_sample_size} evaluates the impact of varying the amount of source domain data on the prediction performance in the target domain. The results indicate that as the volume of source domain data increases, both \method and FTL contribute to improving the prediction accuracy of the target domain. However, \method consistently facilitates the most significant performance enhancement during the knowledge transfer process.

\textbf{Label Heterogeneity}. Among existing methods, VFedTrans and FedHSSL address the issue of label heterogeneity by ensuring that label information from the source domain is not included during knowledge transfer. 
As shown in Tbl.~\ref{tab:cross_label}, \method excels in scenarios involving heterogeneous labels, achieving the best prediction performance without being affected by label heterogeneity. In contrast, methods like FTL, VFL-Infer, and FedCVT struggle to handle this scenario, leading to a marked decline in their prediction accuracy. This demonstrates \method's robustness in challenging cross-domain environments with label heterogeneity.

\begin{table}[H]
\centering
\footnotesize
\caption{Prediction accuracy in scenarios with heterogeneous labels.}
\adjustbox{max width=\linewidth}{
\begin{tabular}{lcc}
    \toprule
    †\textbf{Method} & \textbf{Leukemia} & \textbf{Pneumonia} \\ \midrule
    \method   & $\textbf{0.8048}$ & $\textbf{0.9022}$ \\
    VFedTrans & $0.7924$ & $0.8923$ \\
    Local     & $0.7812$ & $0.8884$ \\
    FTL       & $0.6785$ & $0.7243$ \\
    VFL-Infer & $0.6229$ & $0.7021$ \\
    FedCVT    & $0.7024$ & $0.7435$ \\
    FedHSSL   & $0.7984$ & $0.8965$ \\
    \bottomrule
\end{tabular}
}
\label{tab:cross_label}
\end{table}

\begin{tcolorbox}[left=0mm, right=0mm, top=0mm, bottom=0mm]
\textbf{Answer to \textbf{RQ2}: \method demonstrates superior performance in cross-domain knowledge transfer by effectively addressing both feature heterogeneity and label heterogeneity. 
}
\end{tcolorbox}

\subsection{Few-Shot Evaluation}
\label{subsec:fewshot}
In real-world scenarios, data holders often possess predominantly unlabeled data with access to only a few labeled samples. This is particularly true for rare medical conditions, where the scarcity of labeled instances creates imbalanced datasets with limited training examples. To evaluate performance under these conditions, we simulate a few-shot learning setting where \task has access to only $10\%$ of the original labeled samples for training the downstream model. 
Tbl.~\ref{tab:eval_few_shot} summarizes the results under this constrained setting. The results indicates that the prediction performance of all methods declines to some extent due to the lack of labeled data. However, \method demonstrates a significant advantage, consistently outperforming other methods on seven datasets. This highlights \method's robustness and efficiency in extracting and utilizing knowledge even in few-shot scenarios, making it well-suited for real-world applications with limited labeled data.

\begin{table*}[ht]
\centering
\footnotesize
\caption{Prediction accuracy under few-shot setting.}
\adjustbox{max width=\linewidth}{
\begin{tabular}{lccccccccc}
    \toprule
    \multirow{2}{*}{†\textbf{Method}} & \multirow{2}{*}{\textbf{MIMIC-\Romannum{3}}} & \multirow{2}{*}{\textbf{Cardio}} & \multirow{2}{*}{\textbf{RNA-Seq}} & \multirow{2}{*}{\textbf{Heart}} & \multirow{2}{*}{\textbf{Sepsis}} & \multicolumn{2}{c}{\textbf{Leukemia}} & \multicolumn{2}{c}{\textbf{Pneumonia}} \\ \cmidrule(lr){7-8} \cmidrule(lr){9-10}
    & & & & & & FH & LH & FH & LH \\ \midrule
    \method & $\textbf{0.8023}$ & $\textbf{0.6364}$ & $\textbf{0.8243}$ & $\textbf{0.6526}$ & $\textbf{0.7025}$ & $\textbf{0.7964}$ & $\textbf{0.7969} $ & $\textbf{0.8859}$ & $\textbf{0.8922}$ \\
    Local & $0.7721$ & $0.6022$ & $0.7984$ & $0.5522$ & $0.6884$ & $0.7538$ & $0.7605$ & $0.8678$ & $0.8763$ \\
    VFedTrans & $0.7944$ & $0.6224$ & $0.8122$ & $0.6175$ & $0.6954$ & - & $0.7896$ & - & $0.8805$ \\
    FTL & $0.7956$ & $0.6229$ & $0.8089$ & $0.6324$ & $0.6969$ & $0.7843$ & $0.6243$ & $0.8787$ & $0.6889$ \\
    VFL-Infer & $0.7702$ & $0.6178$ & $0.8024$ & $0.6102$ & $0.6844$ & - & $0.6187$ & - & $0.6774$ \\
    FedCVT & $0.7733$ & $0.6240$ & $0.8075$ & $0.6074$ & $0.6903$ & - & $0.6785$ & - & $0.7320$ \\
    FedHSSL & $0.7924$ & $0.6309$ & $0.8188$ & $0.6420$ & $0.6987$ & - & $0.7920$ & - & $0.8867$ \\
    \bottomrule
\end{tabular}
}
\label{tab:eval_few_shot}
\end{table*}

\begin{tcolorbox}[left=0mm, right=0mm, top=0mm, bottom=0mm]
\textbf{Answer to \textbf{RQ3}: \method excels in few-shot scenarios by consistently outperforming other methods, demonstrating its ability in leveraging limited labeled data for real-world applications.}
\end{tcolorbox}

\subsection{Robustness Evaluation}
\label{subsec:robustness}
Here, we assess the robustness of \method by modifying its configurations, including the FRL and LKT modules, as well as the downstream machine learning methods.

Tbl.~\ref{tab:eval_robust_ml_cross} and \ref{tab:eval_robust_ml_intra} present the results when \method employs different ML models to train the task classifier. 
\method consistently outperforms the baseline across various ML methods, demonstrating the wide-ranging effectiveness of our knowledge transfer approach across different datasets and models. This also highlights the flexibility of \method, which accommodates the use of any classification model. Such flexibility is especially beneficial in practical medical scenarios, as it enables the classifier to better align with the specific characteristics of the target patients.

Additionally, we evaluate the impact of changing the modules in the FRL and LKT modules of \method and VFedTrans. For the FRL module, we experimented with FedSVD~\cite{chai2022practical} and VFedPCA~\cite{cheung2022vertical}. For the LKT module, we used vanilla AE, VAE and WAE. 
Tbl.~\ref{tab:eval_robust_module} summarize the prediction accuracy for different module configurations. The results show that the final accuracy remains robust to these modifications.

\begin{table}[ht]
\centering
\footnotesize
\caption{Cross-domain prediction accuracy under different downstream models.}
\adjustbox{max width=\linewidth}{
\begin{tabular}{lcccc}
    \toprule
    \multirow{2}{*}{†\textbf{Method}} & \multicolumn{2}{c}{\textbf{Leukemia}} & \multicolumn{2}{c}{\textbf{Pneumonia}} \\ \cmidrule(lr){2-3} \cmidrule(lr){4-5}
    & FH & LH & FH & LH \\ \midrule
    \method (CNN)   & $0.8030$ & $0.8048$ & $\textbf{0.9023}$ & $\textbf{0.9022}$ \\
    VFedTrans (CNN) & - & $0.7924$ & - & $0.8923$ \\
    Local  (CNN) & $0.7756$ & $0.7812$ & $0.8903$ & $0.8884$ \\ \midrule
    \method (VGG16)   & $\textbf{0.8053}$ & $\textbf{0.8064}$ & $0.8995$ & $0.8981$ \\
    VFedTrans (VGG16) & - & $0.7904$ & - & $0.8867$ \\
    Local  (VGG16) & $0.7812$ & $0.7803$ & $0.8885$ & $0.8892$ \\ \midrule
    FTL       & $0.7864$ & $0.6785$ & $0.8955$ & $0.7243$ \\
    VFL-Infer & - & $0.6229$ & - & $0.7021$ \\
    FedCVT    & - & $0.7024$ & - & $0.7435$ \\
    FedHSSL   & - & $0.7984$ & - & $0.8965$ \\
    \bottomrule
\end{tabular}
}
\label{tab:eval_robust_ml_cross}
\end{table}

\begin{table}[ht]
\centering
\footnotesize
\caption{Intra-domain prediction accuracy under different downstream models.}
\adjustbox{max width=\linewidth}{
\begin{tabular}{lccccc}
    \toprule
    †\textbf{Method} & \textbf{MIMIC-\Romannum{3}} & \textbf{Cardio} & \textbf{RNA-Seq} & \textbf{Heart} & \textbf{Sepsis} \\ \midrule
    \method (XGBoost) & $\textbf{0.8080}$ & $0.7132$ & $0.9875$ & $\textbf{0.6620}$ & $\textbf{0.8425}$ \\
    VFedTrans (XGBoost) & $0.7983$ & $0.6777$ & $0.9720$ & $0.6425$ & $0.8310$ \\
    Local (XGBoost) & $0.7724$ & $0.6632$ & $0.9250$ & $0.5828$ & $0.7903$ \\ \midrule
    \method (AdaBoost) & $0.7942$ & $\textbf{0.7150}$ & $0.9820$ & $0.6589$ & $0.8386$ \\
    VFedTrans (AdaBoost) & $0.7785$ & $0.7065$ & $0.9750$ & $0.6407$ & $0.8295$ \\
    Local (AdaBoost) & $0.7533$ & $0.6604$ & $0.9284$ & $0.5850$ & $0.7854$ \\ \midrule
    \method (FCNN) & $0.8025$ & $0.7035$ & $\textbf{0.9935}$ & $0.6612$ & $0.8395$ \\
    VFedTrans (FCNN) & $0.7910$ & $0.6873$ & $0.9837$ & $0.6427$ & $0.8344$ \\
    Local (FCNN) & $0.7625$ & $0.6578$ & $0.9364$ & $0.5835$ & $0.7866$ \\ \midrule
    \method (TabNet) & $0.7983$ & $0.6972$ & $0.9915$ & $0.6592$ & $0.8403$ \\
    VFedTrans (TabNet) & $0.7830$ & $0.6825$ & $0.9806$ & $0.6388$ & $0.8330$ \\
    Local (TabNet) & $0.7675$ & $0.6535$ & $0.9325$ & $0.5815$ & $0.7890$ \\ \midrule
    FTL & $0.7656$ & $0.6589$ & $0.9520$ & $0.6021$ & $0.7885$ \\
    VFL-Infer & $0.7723$ & $0.6620$ & $0.9500$ & $0.6123$ & $0.8020$ \\
    FedCVT & $0.7754$ & $0.6626$ & $0.9460$ & $0.6223$ & $0.7985$ \\
    FedHSSL & $0.7955$ & $0.6952$ & $0.9650$ & $0.6520$ & $0.8280$ \\
    \bottomrule
\end{tabular}
}
\label{tab:eval_robust_ml_intra}
\end{table}

\begin{table*}[ht]
\centering
\footnotesize
\caption{Prediction accuracy by changing FRL and LKT modules.}
\adjustbox{max width=\linewidth}{
\begin{tabular}{lccccccccccc}
    \toprule
    \multirow{2}{*}{†\textbf{Method}} & \multirow{2}{*}{\textbf{FRL}} & \multirow{2}{*}{\textbf{LKT}} & \multirow{2}{*}{\textbf{MIMIC-\Romannum{3}}} & \multirow{2}{*}{\textbf{Cardio}} & \multirow{2}{*}{\textbf{RNA-Seq}} & \multirow{2}{*}{\textbf{Heart}} & \multirow{2}{*}{\textbf{Sepsis}} & \multicolumn{2}{c}{\textbf{Leukemia}} & \multicolumn{2}{c}{\textbf{Pneumonia}} \\ \cmidrule(lr){9-10} \cmidrule(lr){11-12}
    & & & & & & & & FH & LH & FH & LH \\ \midrule
    \multirow{6}{*}{\method} & FedSVD & AE & $0.8080$ & $\textbf{0.7132}$ & $\textbf{0.9875}$ & $0.6620$ & $0.8425$ & $\textbf{0.8030}$ & $\textbf{0.8048}$ & $\textbf{0.9023}$ & $\textbf{0.9022}$ \\
    & FedSVD & VAE & $0.8065$ & $0.7122$ & $0.9850$ & $\textbf{0.6634}$ & $0.8417$ & $0.8011$ & $0.8019$ & $0.8996$ & $0.9005$ \\
    & FedSVD & WAE & $\textbf{0.8089}$ & $0.7105$ & $0.9870$ & $0.6615$ & $\textbf{0.8432}$ & $0.8025$ & $0.7980$ & $0.9010$ & $0.9015$ \\
    & VFedPCA & AE & $0.8058$ & $0.7129$ & $0.9820$ & $0.6609$ & $0.8388$ & $0.7988$ & $0.7992$ & $0.8978$ & $0.8982$ \\
    & VFedPCA & VAE & $0.8065$ & $0.7095$ & $0.9840$ & $0.6613$ & $0.8397$ & $0.7982$ & $0.7978$ & $0.8984$ & $0.8990$ \\
    & VFedPCA & WAE & $0.8048$ & $0.7102$ & $0.9830$ & $0.6578$ & $0.8390$ & $0.7994$ & $0.8004$ & $0.8988$ & $0.8995$ \\ \midrule
    \multirow{6}{*}{VFedTrans} & FedSVD & AE & $0.7983$ & $0.6777$ & $0.9720$ & $0.6425$ & $0.8310$ & - & $0.7924$ & - & $0.8923$ \\
    & FedSVD & VAE & $0.8005$ & $0.6754$ & $0.9740$ & $0.6402$ & $0.8344$ & - & $0.7912$ & - & $0.8905$ \\
    & FedSVD & WAE & $0.7975$ & $0.6768$ & $0.9710$ & $0.6434$ & $0.8353$ & - & $0.7895$ & - & $0.8912$ \\
    & VFedPCA & AE & $0.7965$ & $0.6745$ & $0.9765$ & $0.6389$ & $0.8295$ & - & $0.7940$ & - & $0.8893$ \\
    & VFedPCA & VAE & $0.7957$ & $0.6712$ & $0.9718$ & $0.6394$ & $0.8288$ & - & $0.7902$ & - & $0.8905$ \\
    & VFedPCA & WAE & $0.7975$ & $0.6726$ & $0.9685$ & $0.6412$ & $0.8308$ & - & $0.7895$ & - & $0.8885$ \\ \midrule
    Local & - & - & $0.7724$ & $0.6632$ & $0.9250$ & $0.5828$ & $0.7903$ & $0.7756$ & $0.7812$ & $0.8903$ & $0.8884$ \\
    FTL & - & - & $0.7656$ & $0.6589$ & $0.9520$ & $0.6021$ & $0.7885$ & $0.7864$ & $0.6785$ & $0.8955$ & $0.7243$ \\
    VFL-Infer & - & - & $0.7723$ & $0.6620$ & $0.9500$ & $0.6123$ & $0.8020$ & - & $0.6229$ & - & $0.7021$ \\
    FedCVT & - & - & $0.7754$ & $0.6626$ & $0.9460$ & $0.6223$ & $0.7985$ & - & $0.7024$ & - & $0.7435$ \\
    FedHSSL  & - & - & $0.7955$ & $0.6952$ & $0.9650$ & $0.6520$ & $0.8280$ & - & $0.7984$ & - & $0.8965$ \\
    \bottomrule
\end{tabular}
}
\label{tab:eval_robust_module}
\end{table*}

\begin{tcolorbox}[left=0mm, right=0mm, top=0mm, bottom=0mm]
\textbf{Answer to \textbf{RQ4}: \method demonstrates exceptional robustness and flexibility, consistently outperforming baselines across diverse ML models, datasets, and module configurations, highlighting its adaptability and effectiveness in practical scenarios such as medical applications.}
\end{tcolorbox}

\subsection{Scalability Evaluation}
\label{subsec:eval_scalability}
In real-world scenarios, \task may collaborate with multiple $\mathcal{P}_d$s to enhance local medical prediction services. As the number of $\mathcal{P}_d$s increases, the knowledge transfer method must satisfy two key conditions: 
\ding{172} Computational Efficiency: The process of transferring knowledge from each \data must remain computationally efficient. 
\ding{173} Knowledge Integration: It must effectively acquire knowledge from different $\mathcal{P}_d$s.

\textbf{Computational Efficiency}. The computational burden of \method primarily depends on the FRL module, as this is the only step that involves collaboration between \task and \data. It is important to note that our implemented FRL algorithm, such as FedSVD, is highly efficient and can be applied to feature matrices at a billion-scale~\cite{chai2022practical}. This significantly contributes to the overall computational efficiency of \method. 
We first evaluate the computational efficiency by varying the sample size of $H_t^{ol}$ on MIMIC-\Romannum{3} and Cardio datasets. Using the sample size in the default experimental settings as the lower limit (100\%), we test how \method performs with sample sizes ranging from 100\% to 500\%. The left graph of Fig.~\ref{fig:scala_time} illustrates that as the size of overlapping samples increases, the computational overhead for \method remains low and shows only a minimal increase. 
Next, we assess the computational efficiency of \method under scenarios involving multiple $\mathcal{P}_d$s. Unlike HFL, the number of participants in VFL is typically limited to single digits. As shown in the right graph of Fig.~\ref{fig:scala_time}, the computational overhead of \method increases linearly with the number of $\mathcal{P}_d$s. This linear relationship demonstrates the scalability and efficiency of our mechanism in handling additional participants.

\textbf{Knowledge Integration}. As the number of $\mathcal{P}_d$s increases, effectively transferring knowledge from multiple sources to $H_t^{nl}$ becomes a significant challenge. To evaluate the ability of \method to aggregate multi-party knowledge, we vary the number of $\mathcal{P}_d$s from 1 to 9 with a step size of 2. The evaluation results are shown in Fig.~\ref{fig:scala_data_hospital_accuracy}. 
Among the existing methods, FTL and FedCVT are limited to 2-party scenarios, which restricts their applicability in real-world settings. As the number of $\mathcal{P}_d$s increases, the marginal improvement in predictive performance from richer knowledge becomes limited. Methods such as VFedTrans, VFL-Infer, and FedHSSL experience a decrease in prediction accuracy under these conditions. In contrast, \method is able to learn non-redundant knowledge, allowing it to maintain high prediction accuracy even with an increasing number of data hospitals.

\begin{tcolorbox}[left=0mm, right=0mm, top=0mm, bottom=0mm]
\textbf{Answer to \textbf{RQ5}: Our approach, \method, demonstrates superior scalability by maintaining efficient knowledge transfer handling multiple $\mathcal{P}_d$s, ensuring high prediction accuracy even as the number of $\mathcal{P}_d$s grows.}
\end{tcolorbox}

\subsection{Ablation Study}

To further validate the effectiveness of \method in knowledge transfer, we compare the changes in prediction accuracy before and after incorporating $\mathcal{L}_{mi}$. As shown in Fig.~\ref{fig:ablation_mi}, the prediction accuracy improves significantly when $\mathcal{L}_{mi}$ is used for knowledge transfer. Similarly, when the features of \task change, our mechanism consistently outperforms the version without $\mathcal{L}_{mi}$. This demonstrates that \method enables hospitals to greatly benefit from cross-institution collaboration.

We then evaluated the role of $\mathcal{L}_{cl}$ in scenarios involving more $\mathcal{P}_d$s. As shown in Fig.~\ref{fig:ablation_cl}, $\mathcal{L}_{cl}$ effectively eliminates redundant information, enabling the prediction performance of \task to improve slightly as the number of $\mathcal{P}_d$s increases. This demonstrates
\method's ability to enhance scalability and maintain efficient knowledge transfer in multi-party collaboration.

\section{Conclusion}
In this study, we present a unified vertical federated knowledge transfer framework (\method) designed to transfer knowledge from cross-party overlapping samples to each hospital's local non-overlapping samples in a domain-adaptive manner. \method exhibits robust dual capabilities, excelling in both intra-domain and cross-domain knowledge transfer. Comprehensive experiments conducted on seven medical datasets validate the framework's effectiveness in enhancing medical prediction performance for non-overlapping, vulnerable patient groups within participating hospitals.

\ifCLASSOPTIONcaptionsoff
  \newpage
\fi



\bibliographystyle{IEEEtran}
\bibliography{reference}
%



%
\vspace{-1cm}
\begin{IEEEbiography}[{\includegraphics[width=1in,height=1.25in,clip,keepaspectratio]{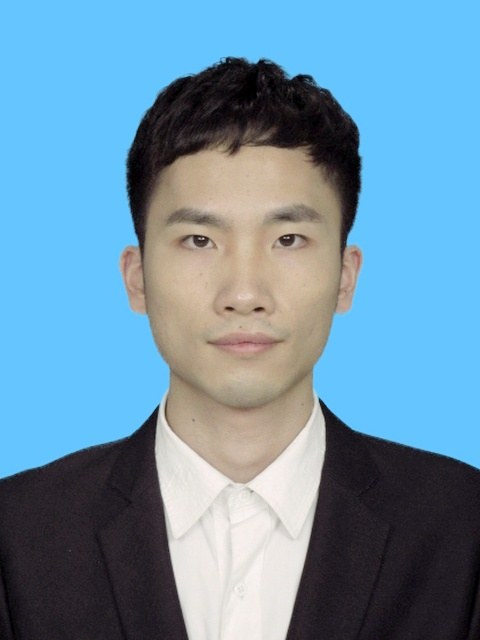}}]{Chung-ju Huang} is a Ph.D student in Key Laboratory of High Confidence Software Technologies (Peking University), Ministry of Education, Beijing, 100871, China; School of Computer Science, Peking University, Beijing, 100871, China. His interests include federated learning and responsible AI.
\end{IEEEbiography}
\vspace{-0.5cm}
\begin{IEEEbiography}[{\includegraphics[width=1in,height=1.25in,clip,keepaspectratio]{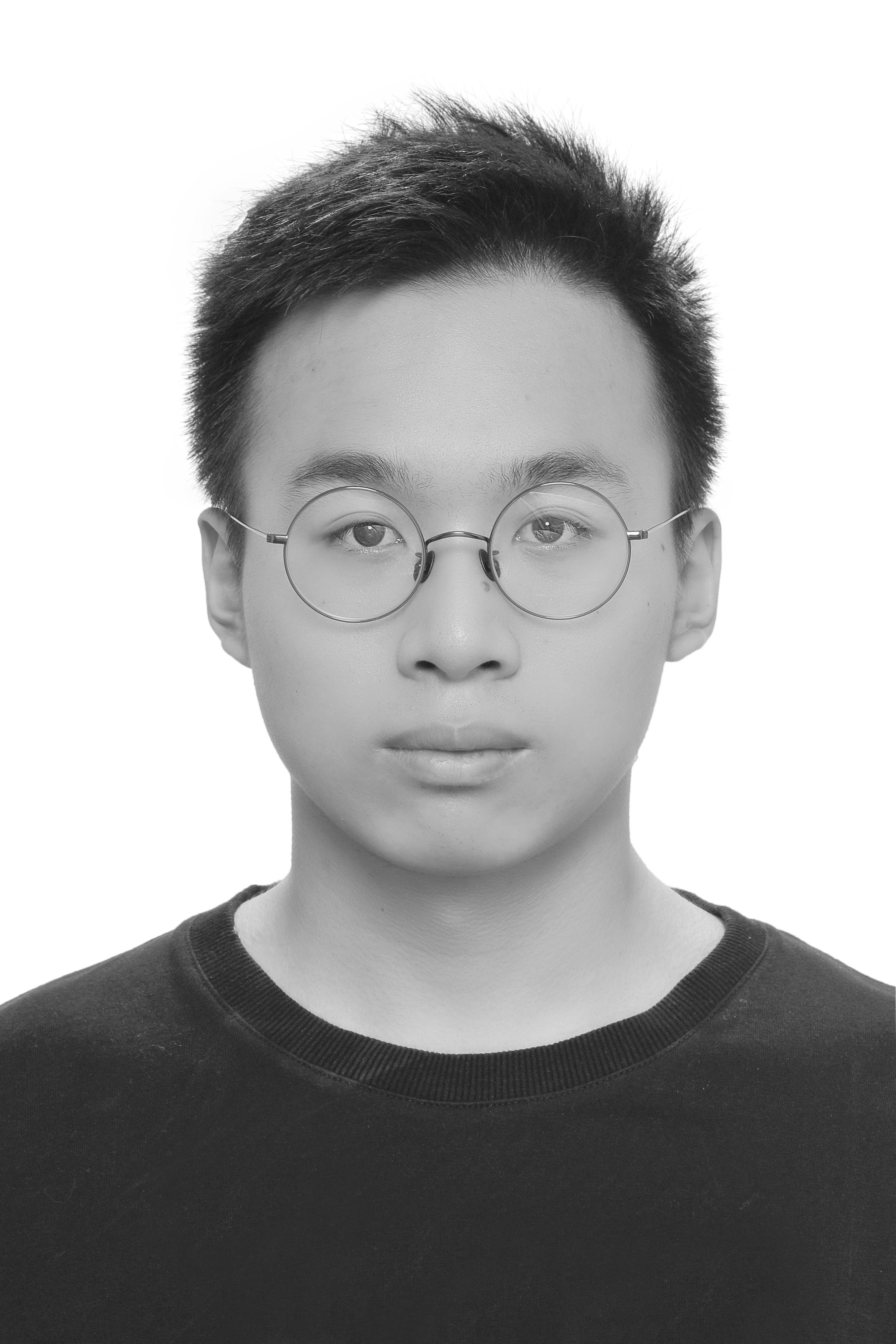}}]{Yuanpeng He} is a Ph.D student in Key Laboratory of High Confidence Software Technologies (Peking University), Ministry of Education, Beijing, 100871, China; School of Computer Science, Peking University, Beijing, 100871, China. His interests include computer vision and adaptive software engineering.
\end{IEEEbiography}
\vspace{-0.5cm}
\begin{IEEEbiography}[{\includegraphics[width=1in,height=1.25in,clip,keepaspectratio]{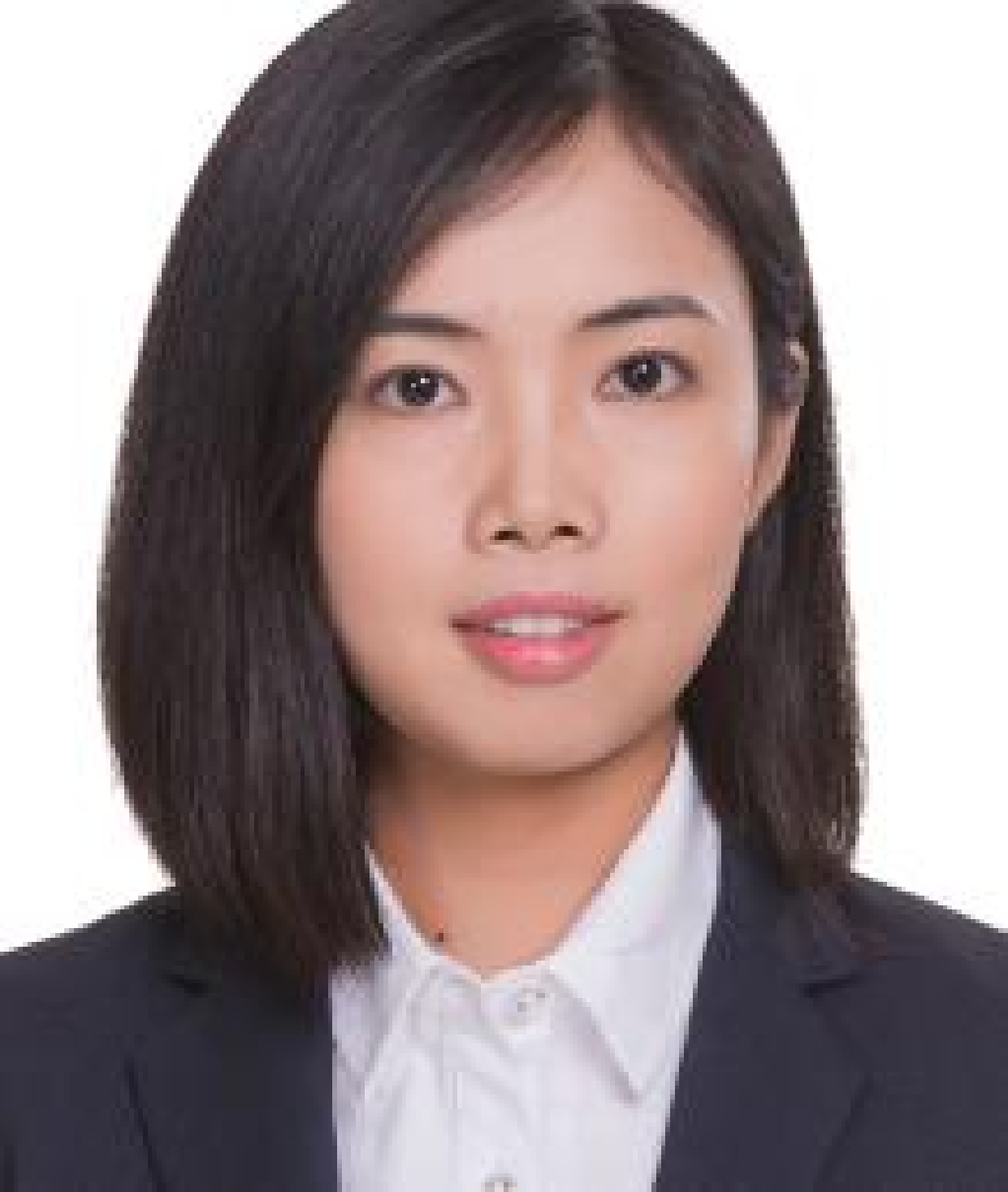}}]{Xiao Han} received the Ph.D. degree in computer science from Pierre and Marie Curie University and the Institut Mines-TELECOM/TELECOM SudParis in 2015. She is currently a Full Professor with the Beihang University, China. Her research interests include social network analysis, fintech, and privacy protection.
\end{IEEEbiography}
\vspace{-1cm}
\begin{IEEEbiography}[{\includegraphics[width=1in,height=1.25in,clip,keepaspectratio]{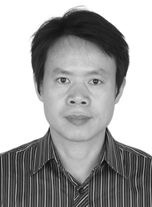}}]{Wenpin Jiao} is a professor in the Department of Computer Science and technology, School of EECS, Peking University. He obtained his B.S. and M.S. from East China University of Science and Technology in 1991 and 1997, respectively, and Ph.D. from Institute of Software, Chinese Academy of Sciences in 2000. His research interests include software engineering, adaptive software, multi-agent systems.

Prof. Jiao has published more than 70 research papers, and many of them are published in reputable international journals and conferences, such as Journal of Software ans Systems, IET Software, Computational Intelligence, FSE, ICSR. He is serving as a member of CAAI Knowledge Engineering and Distributed Intelligence Specialized Committee.
\end{IEEEbiography}
\vspace{-1cm}
\begin{IEEEbiography}[{\includegraphics[width=1in,height=1.25in,clip,keepaspectratio]{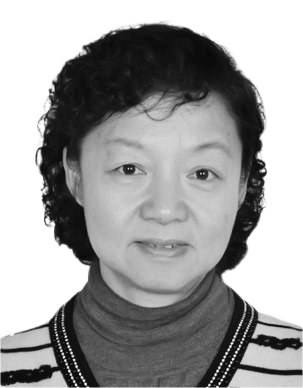}}]{Zhi Jin (Fellow, IEEE)} received the PhD degree in computer science from Changsha Institute of Technology, China, in 1992. She is currently a professor of computer science at Peking University and the deputy director of Key Lab of High Confidence Software Technologies (Ministry of Education) at Peking University. Her research interests include software engineering, requirements Engineering, knowledge engineering, and machine learning. She has published over 200 scientific articles in refereed international journals, such as IEEE T-KDE, T-SE, T-R, T-ITS, T-PDS, ACM T-OSEM, and T-CPS. She has held more than 30 approved invention patents.

She has coauthored five books. She is/was principal investigator of more than 15 national competitive grants. Prof. Jin served/serves as an Associate Editor of IEEE Transactions on Software Engineering, IEEE Transactions on Reliability, and ACM Transactions on Autonomous and Adaptive Systems. She served/serves on the Editorial Board of Requirements Engineering Journal and Empirical Software Engineering. She is currently a Fellow of the IEEE, Fellow of CCF (China Computer Federation), and the director of CCF Technical Committee of System Software.
\end{IEEEbiography}
\vspace{-1cm}
\begin{IEEEbiography}[{\includegraphics[width=1in,height=1.25in,clip,keepaspectratio]{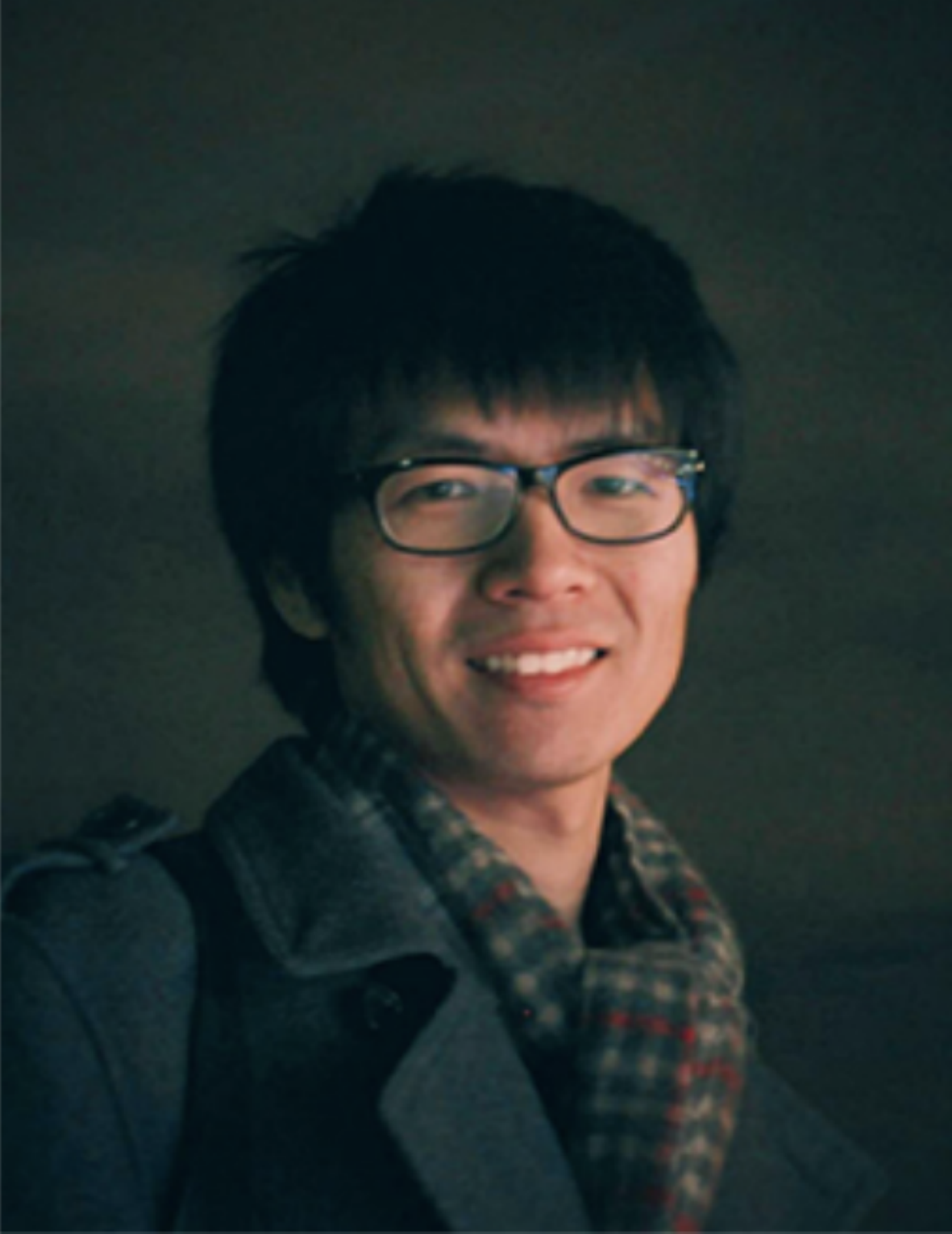}}]{Leye Wang} is a tenured associate professor at Key Lab of High Confidence Software Technologies (Peking University), MOE, and School of Computer Science, Peking University, China. He received a Ph.D. in computer science from TELECOM SudParis and University Paris 6, France, in 2016. He was a postdoc researcher with Hong Kong University of Science and Technology. His research interests include ubiquitous computing, mobile crowdsensing, and urban computing.
\end{IEEEbiography}







\newpage
\appendices

\section{Problem Formulation}

We list the mathematical notions used in this paper in Tbl.~\ref{tab:notion}.

\begin{table}[htbp]
\centering
\footnotesize
\caption{List of used notions.}
\begin{tabular}{@{}ll@{}}
    \toprule
    \ding{61}\textbf{Notation} & \textbf{Description} \\ \midrule
    \textbf{\task,\data} & Task hospital and Data hospital. \\
    $H_t^{ol},H_d^{ol}$ & Overlapping patients between \task and \data \\
    $H_t^{nl},H_d^{nl}$ & Non-overlapping patients \\
    $X_t,X_d$ & Feature space of \task and \data. \\
    $I_t,I_d$ & Patient space of \task and \data. \\
    $Y_t$ & Labels of \task. \\
    $H_{fed}^{ol}$ & Federated latent representations. \\
    $Enc$ & Encoder in LKT module. \\
    $\mathcal{L}_{recons}$ & Reconstruction loss. \\
    $\mathcal{L}_{mi}$ & Mutual information loss. \\
    $\mathcal{L}_{cl}$ & Contrastive loss. \\
    \bottomrule
\end{tabular}
\label{tab:notion}
\end{table}

\section{Evaluation Setup}
\label{app:evaluation_setup}
\begin{table}[htbp]
\centering
\footnotesize
\caption{Evaluation Datasets.}
\adjustbox{max width=\linewidth}{
\begin{tabular}{cccc}
    \toprule
    \textbf{†Setting} &\textbf{Dataset} & \textbf{Dimension} & \textbf{Class} \\ \midrule
    \multirow{5}{*}{Intra-domain} & MIMIC-\Romannum{3}~\cite{johnson2016mimic} & $58,976\times15$ & 4 \\
    & Cardio~\cite{cardiovascular} & $70,000\times 11$ & 2 \\
    & RNA-Seq~\cite{weinstein2013cancer} & $801 \times 20,531$ & 5 \\
    & Heart~\cite{heart-attack} & $8,763\times 26$ & 2 \\
    & Sepsis~\cite{sepsis} & $45,852\times 17$ & 2 \\ \midrule
    \multirow{2}{*}{Cross-domain} & Leukemia & $12,528\times 3\times 128\times 128$ & 2 \\
    & Pneumonia & $5,863\times 3\times 64 \times 64$ & 2 \\
    \bottomrule
\end{tabular}
}
\label{tab:dataset}
\end{table}

\begin{table}[htbp]
\centering
\footnotesize
\caption{Default data partition.}
\adjustbox{max width=\linewidth}{
\begin{tabular}{@{}cccccc@{}}
    \toprule
    \multirow{2}{*}{†\textbf{Dataset}} & \multirow{2}{*}{$H_t^{nl}$} & \multicolumn{2}{c}{\textbf{Single Mode}} & \multicolumn{2}{c}{\textbf{Multiple Mode}} \\ \cmidrule(lr){3-4} \cmidrule(lr){5-6}
    & & $H_t^{ol}$ & $H_d^{ol}$ & $H_t^{ol}$ & $H_d^{ol}$ \\ \midrule
    MIMIC-\Romannum{3} & $30,000\times5$ & $10,000\times5$ & $10,000\times5$ & $1,000\times5$ & $1,000\times5$ \\
    Cardio & $30,000\times5$ & $10,000\times5$ & $10,000\times5$ & $1,000\times5$ & $1,000\times5$ \\
    RNA-Seq & $400\times 5,000$ & $200\times 5,000$ & $200\times 5,000$ & $20\times 5,000$ & $20\times 5,000$ \\
    Heart & $3000\times 9$ & $2000\times 9$ & $2000\times 5$ & $200\times 9$ & $200\times 5$\\ 
    Sepsis & $30,000\times5$ & $10,000\times5$ & $10,000\times5$ & $1,000\times5$ & $1,000\times5$ \\ \midrule
    \multirow{2}{*}{Leukemia} & $5000\times$ & \multicolumn{2}{c}{$1000\times$}  & \multicolumn{2}{c}{$200\times$} \\
    & $3\times 32\times 32$ & \multicolumn{2}{c}{$3\times 128\times 64$} & \multicolumn{2}{c}{$3\times 128\times 64$} \\
    \multirow{2}{*}{Pneumonia} & $3000\times$ & \multicolumn{2}{c}{$1000\times$}  & \multicolumn{2}{c}{$200\times$} \\
    & $3\times 24\times 24$ & \multicolumn{2}{c}{$3\times 64\times 32$} & \multicolumn{2}{c}{$3\times 64\times 32$} \\
    \bottomrule
\end{tabular}
}
\label{tab:data_split}
\end{table}

\begin{table}[htbp]
\centering
\footnotesize
\caption{Default key parameters in LKT modules.}
\adjustbox{max width=\linewidth}{
\begin{tabular}{cccc}
    \toprule
    †\textbf{Backbone} &\textbf{Parameter} & \textbf{Value} & \textbf{Description} \\ \midrule
    \multirow{2}{*}{AE} & depth & 3 & The depth of encoder and decoder. \\
    & neuron & Sigmoid & The activation function of hidden layers. \\ \midrule
    \multirow{2}{*}{VAE} & depth & 2 & The depth of encoder and decoder. \\
    & kld\_weight & 0.5 & The weight of $\mathcal{L}_{KL}$. \\ \midrule
    \multirow{2}{*}{WAE} & d\_depth & 4 & The depth of discriminator. \\
    & depth & 2 & The depth of encoder and decoder. \\
    \bottomrule
\end{tabular}
}
\label{tab:lrd_setting}
\end{table}

\begin{table}[htbp]
\centering
\footnotesize
\caption{Default key parameters in task-specific ML models.}
\adjustbox{max width=\linewidth}{
\begin{tabular}{ccc}
    \toprule
    †\textbf{Model} & \textbf{Parameter} & \textbf{Default} \\ \midrule
    \multirow{2}{*}{MLP} & Hidden layer & $4\times$ FCNN \\ 
    & Activation function & ReLU \\ \midrule
    \multirow{3}{*}{TabNet} & Decision features & 32 \\ 
    & Attention features & 32  \\ 
    & Decision steps & 3 \\ \midrule
    \multirow{2}{*}{CNN} & Hidden layer & 4$\times$Conv+2$\times$FCNN \\ 
    & Activation function & Relu \\ 
    \midrule
    \multirow{2}{*}{VGG16} & Hidden layer & 13$\times$Conv+2$\times$FCNN \\ 
    & Activation function & Relu \\ 
    \bottomrule
\end{tabular}
}
\label{tab:ml_setting}
\end{table}

\end{document}